%% file: root.tex
\documentclass[letterpaper, 10 pt, journal, twoside]{IEEEtran}

\IEEEoverridecommandlockouts                              %

\usepackage{graphicx} %
\usepackage{amsmath} %
\usepackage{amssymb}  %
\usepackage[caption=false, font=footnotesize]{subfig}
\usepackage{csvsimple}
\usepackage{tikz} 
\usetikzlibrary{positioning}
\usepackage{hyperref} %

\title{%
Hearing What You Cannot See:\\ Acoustic Vehicle Detection Around Corners
}

\author{Yannick Schulz$^{*1}$, Avinash Kini Mattar$^{*1}$, Thomas M. Hehn$^{*1}$, and Julian F. P. Kooij$^1$%
\thanks{Manuscript received: October, 15, 2020; Revised January, 9, 2021;
Accepted February, 7, 2021.}%
\thanks{This paper was recommended for publication by Editor Cesar Cadena
upon evaluation of the Associate Editor and Reviewers' comments.}%
\thanks{*) Shared first authors. 1) Intelligent Vehicles Group, TU Delft, The Netherlands. Primary contact:
        {\tt\small J.F.P.Kooij@tudelft.nl}}%
\thanks{Digital Object Identifier (DOI): 10.1109/LRA.2021.3062254}
}

\newcommand\copyrighttext{%
    \footnotesize \textcopyright 2021 IEEE. Personal use of this material is permitted.
    Permission from IEEE must be obtained for all other uses, in any current or future
    media, including reprinting/republishing this material for advertising or promotional
    purposes, creating new collective works, for resale or redistribution to servers or
    lists, or reuse of any copyrighted component of this work in other works.
    DOI: 10.1109/LRA.2021.3062254}
\newcommand\copyrightnotice{%
    \begin{tikzpicture}[remember picture,overlay]
    \node[anchor=south,yshift=0pt] at (current page.south) {\fbox{\parbox{\dimexpr\textwidth-\fboxsep-\fboxrule\relax}{\copyrighttext}}};
    \end{tikzpicture}%
}

\DeclareMathOperator*{\argmax}{arg\,max\,}

\begin{document}

\maketitle

\newcommand{\Updated}[1]{{#1}} %

\newcommand{\classleft}{{\tt left}}
\newcommand{\classright}{{\tt right}}
\newcommand{\classfront}{{\tt front}}
\newcommand{\classneg}{{\tt none}}
\newcommand{\classset}{\mathcal{C}} %
\newcommand{\classifier}{f} 
\newcommand{\nostaticrecordings}{623}
\newcommand{\nodrivingrecordings}{441} %

\newcommand{\samplelen}{\delta t} %
\newcommand{\minfreq}{f_{min}} %
\newcommand{\maxfreq}{f_{max}}
\newcommand{\numsegments}{L} %
\newcommand{\ypred}{y^*} %
\newcommand{\alphathreshold}{\alpha_{\mathrm{th}}} %
\newcommand{\ytrue}{\overline{y}} %
\newcommand{\nummics}{M} %
\newcommand{\numtestdata}{N} %

\newcommand{\singleplotxy}[2]{
         \begin{tikzpicture}
	     \node[anchor=south west,inner sep=0] (image2) at (0,0) 
	     {\includegraphics[width=0.295\textwidth]{#2}};
	     \node[inner sep=0] at (3.0,-0.1) {$t_e-#1_0$ [s]};
	     \node[inner sep=0,rotate=90,left=of image2,yshift=-0.8cm,xshift=0.6cm]
	     {$p(c|\boldsymbol{x})$};
	     \end{tikzpicture}}
\newcommand{\singleplotx}[2]{
         \begin{tikzpicture}
	     \node[anchor=south west,inner sep=0] (image1) at (0,0) 
	     {\includegraphics[width=0.295\textwidth]{#2}};
	     \node[inner sep=0] at (3.0,-0.1) {$t_e-#1_0$ [s]};
	     \end{tikzpicture}}
\newcommand{\doubleplotxy}[5]{
         \begin{tikzpicture}
	     \node[anchor=south west,inner sep=0] (image2) at (0,0) 
	     {\includegraphics[width=0.295\textwidth]{#4}};
	     \node[inner sep=0] at (3.0,-0.1) {$t_e-#1_0$ [s]};
	     \node[above=of image2,anchor=center,inner sep=0,yshift=0.4cm] (image1)%
	     {\includegraphics[width=0.295\textwidth]{#2}};
	     \node[inner sep=0,rotate=90,left=of image1,anchor=center,yshift=-0.3cm,outer sep=0cm] %
	     {#3};
	     \node[inner sep=0,rotate=90,left=of image2,anchor=center,yshift=-0.3cm,outer sep=0cm] %
	     {#5};
	     \node[inner sep=0,rotate=90,left=of image2,yshift=-0.8cm,xshift=0.6cm]
	     {$p(c|\boldsymbol{x})$};
	     \node[inner sep=0,rotate=90,left=of image1,yshift=-0.8cm,xshift=0.6cm]
	     {$p(c|\boldsymbol{x})$};
	     \end{tikzpicture}}
\newcommand{\doubleplotx}[3]{
         \begin{tikzpicture}
	     \node[anchor=south west,inner sep=0] (image1) at (0,0) 
	     {\includegraphics[width=0.295\textwidth]{#3}};
	     \node[inner sep=0] at (3.0,-0.1) {$t_e-#1_0$ [s]};
	     \node[above=of image1,anchor=center,inner sep=0,yshift=0.4cm] (image2)%
	     {\includegraphics[width=0.295\textwidth]{#2}};
	     \end{tikzpicture}}
	     
\newcommand{\trippleplotxy}[6]{
         \begin{tikzpicture}
	     \node[anchor=south west,inner sep=0] (image1) at (0,0) 
	     {\includegraphics[width=0.295\textwidth]{#6}};
	     \node[inner sep=0] at (3.0,-0.1) {$t_e-#1_0$ [s]};
	     \node[above=of image1,anchor=center,inner sep=0,yshift=0.4cm] (image2)%
	     {\includegraphics[width=0.295\textwidth]{#4}};
	     \node[above=of image2,anchor=center,inner sep=0,yshift=0.4cm] (image3)%
	     {\includegraphics[width=0.295\textwidth]{#2}};
	     \node[inner sep=0,rotate=90,left=of image3,anchor=center,yshift=-0.3cm,outer sep=0cm] %
	     {#3};
	     \node[inner sep=0,rotate=90,left=of image2,anchor=center,yshift=-0.3cm,outer sep=0cm] %
	     {#5};
	     \node[inner sep=0,rotate=90,left=of image2,yshift=-0.8cm,xshift=0.6cm]
	     {$p(c|\boldsymbol{x})$};
	     \node[inner sep=0,rotate=90,left=of image3,yshift=-0.8cm,xshift=0.6cm]
	     {$p(c|\boldsymbol{x})$};
	     \end{tikzpicture}}
\newcommand{\trippleplotx}[4]{
         \begin{tikzpicture}
	     \node[anchor=south west,inner sep=0] (image1) at (0,0) 
	     {\includegraphics[width=0.295\textwidth]{#4}};
	     \node[inner sep=0] at (3.0,-0.1) {$t_e-#1_0$ [s]};
	     \node[above=of image1,anchor=center,inner sep=0,yshift=0.4cm] (image2)%
	     {\includegraphics[width=0.295\textwidth]{#3}};
	     \node[above=of image2,anchor=center,inner sep=0,yshift=0.4cm] (image3)%
	     {\includegraphics[width=0.295\textwidth]{#2}};
	     \end{tikzpicture}}
	     
\newcommand{\quadrupleplotxy}[7]{
         \begin{tikzpicture}
	     \node[anchor=south west,inner sep=0] (image1) at (0,0) 
	     {\includegraphics[width=0.295\textwidth]{#7}};
	     \node[inner sep=0] at (3.0,-0.1) {$t_e-#1_0$ [s]};
	     \node[above=of image1,anchor=center,inner sep=0,yshift=0.4cm] (image2)%
	     {\includegraphics[width=0.295\textwidth]{#6}};
	     \node[above=of image2,anchor=center,inner sep=0,yshift=0.4cm] (image3)%
	     {\includegraphics[width=0.295\textwidth]{#4}};
	     \node[above=of image3,anchor=center,inner sep=0,yshift=0.4cm] (image4)%
	     {\includegraphics[width=0.295\textwidth]{#2}};
	     \node[inner sep=0,rotate=90,left=of image4,anchor=center,yshift=-0.3cm,outer sep=0cm] %
	     {#3};
	     \node[inner sep=0,rotate=90,left=of image3,anchor=center,yshift=-0.3cm,outer sep=0cm] %
	     {#5};
	     \node[inner sep=0,rotate=90,left=of image2,anchor=center,yshift=-0.3cm,outer sep=0cm] %
	     {#3};
	     \node[inner sep=0,rotate=90,left=of image1,anchor=center,yshift=-0.3cm,outer sep=0cm] %
	     {#5};
	     \node[inner sep=0,rotate=90,left=of image1,yshift=-0.8cm,xshift=0.6cm]
	     {\textit{Recall}};
	     \node[inner sep=0,rotate=90,left=of image2,yshift=-0.8cm,xshift=0.6cm]
	     {\textit{Recall}};
	     \node[inner sep=0,rotate=90,left=of image3,yshift=-0.8cm,xshift=0.6cm]
	     {$p(c|\boldsymbol{x})$};
	     \node[inner sep=0,rotate=90,left=of image4,yshift=-0.8cm,xshift=0.6cm]
	     {$p(c|\boldsymbol{x})$};
	     \end{tikzpicture}}
\newcommand{\quadrupleplotx}[5]{
         \begin{tikzpicture}
	     \node[anchor=south west,inner sep=0] (image1) at (0,0) 
	     {\includegraphics[width=0.295\textwidth]{#5}};
	     \node[inner sep=0] at (3.0,-0.1) {$t_e-#1_0$ [s]};
	     \node[above=of image1,anchor=center,inner sep=0,yshift=0.4cm] (image2)%
	     {\includegraphics[width=0.295\textwidth]{#4}};
	     \node[above=of image2,anchor=center,inner sep=0,yshift=0.4cm] (image3)%
	     {\includegraphics[width=0.295\textwidth]{#3}};
	     \node[above=of image3,anchor=center,inner sep=0,yshift=0.4cm] (image4)%
	     {\includegraphics[width=0.295\textwidth]{#2}};
	     \end{tikzpicture}}
\begin{abstract}

    \input{tex/abstract.tex}

\end{abstract}

\copyrightnotice{}


\section{INTRODUCTION}
    \input{tex/introduction.tex}

\section{Related works}
    \input{tex/relatedworks.tex}

\section{Approach}
    \input{tex/methods.tex}

\section{Experiments}
    \input{tex/experiments.tex}

\section{Conclusions}
    \input{tex/conclusions.tex}

\addtolength{\textheight}{-12cm}   %

\bibliographystyle{IEEEtran}
\bibliography{references}

\end{document}

%% file: tex/abstract.tex
This work proposes to use passive acoustic perception 
as an additional sensing modality for intelligent vehicles.
We demonstrate that approaching vehicles behind blind corners can be detected
by sound before such vehicles enter in line-of-sight.
We have equipped a research vehicle with a roof-mounted microphone array,
and show on data collected with this sensor setup that wall reflections provide information
on the presence and direction of occluded approaching vehicles.
A novel method is presented to classify if and from what direction a vehicle is approaching before it is visible, using as input Direction-of-Arrival features
that can be efficiently computed from the streaming microphone array data.
Since the local geometry around the ego-vehicle affects the perceived patterns,
we systematically study several environment types,
and investigate generalization across these environments.
With a static ego-vehicle, an accuracy of 0.92 is achieved
on the hidden vehicle classification task.
Compared to a state-of-the-art visual detector, Faster R-CNN,
our pipeline achieves the same accuracy more than one second ahead,
providing crucial reaction time for the situations we study.
While the ego-vehicle is driving, we demonstrate positive results on acoustic detection,
still achieving an accuracy of 0.84 within one environment type.
We further study failure cases across environments to identify future research directions.

%% file: tex/introduction.tex
\begin{figure}[t]
    \centering
    \subfloat[line-of-sight sensing]
    {\includegraphics[width=0.48\linewidth] {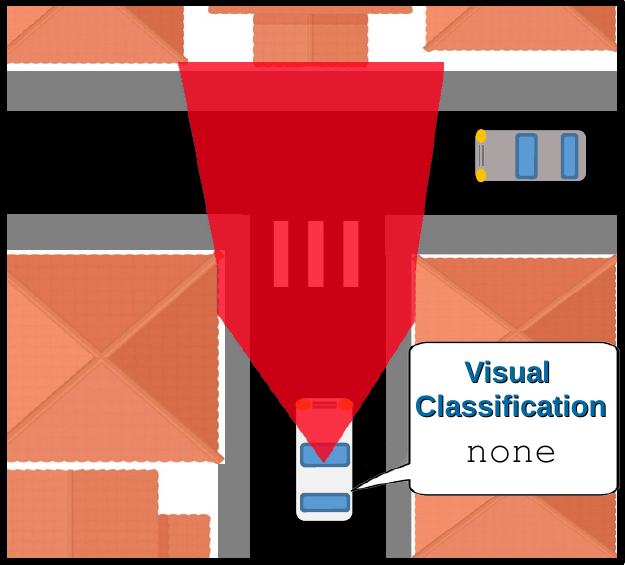}
    \label{fig:schematic-line-of-sight-sensing}
    }
    \subfloat[directional acoustic sensing]
    {\includegraphics[width=0.48\linewidth] {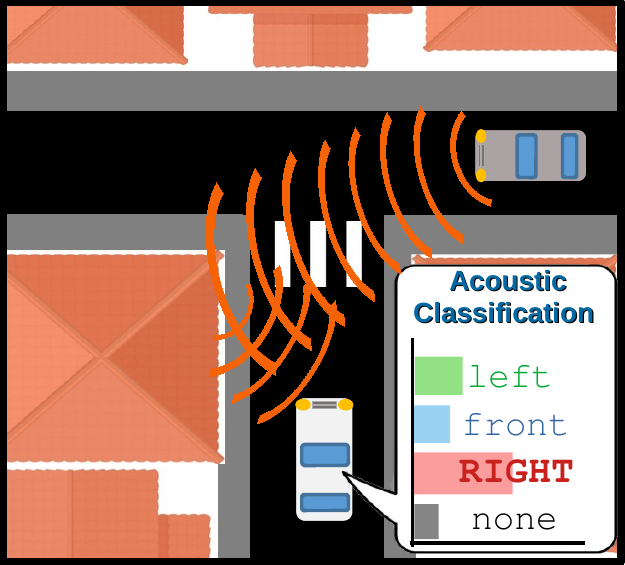}
    \label{fig:schematic-acoustic-sensing}
    }
    \\
    \subfloat[sound localization with a vehicle-mounted microphone array detects the
    wall reflection of an approaching vehicle behind a corner before it appears]
    {%
    \includegraphics[width=0.48\linewidth] {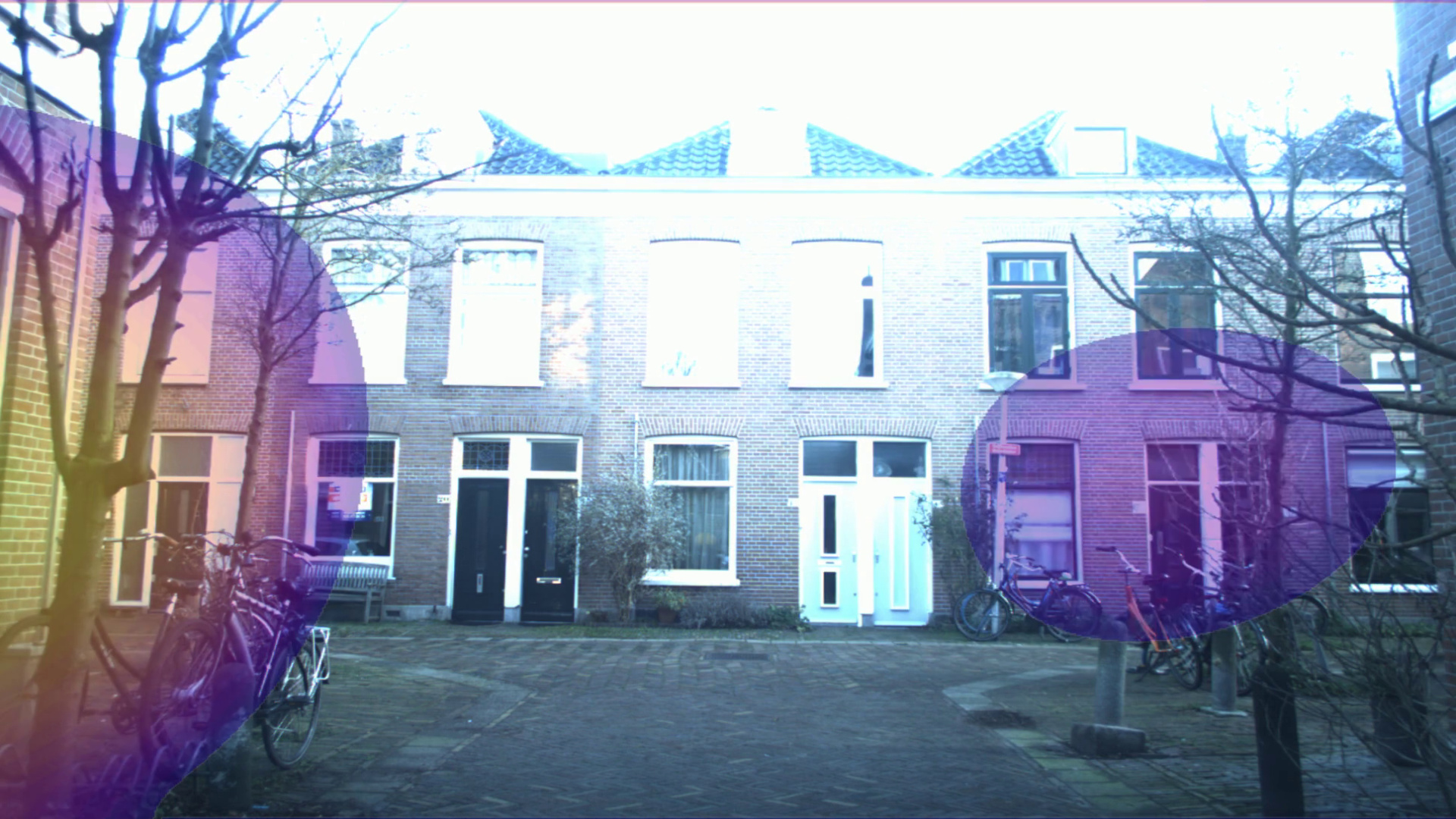}
    \includegraphics[width=0.48\linewidth] {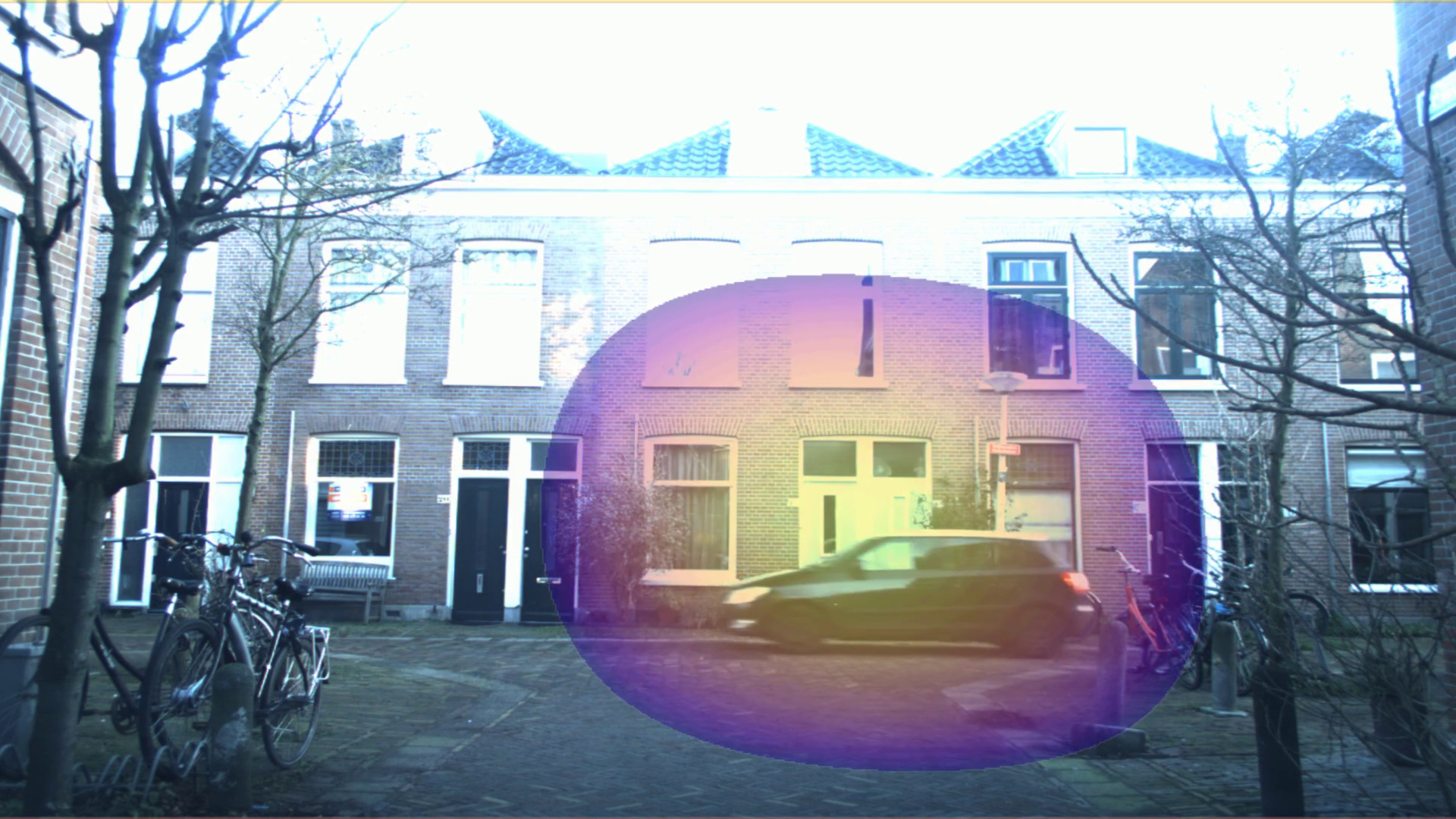}
    \label{fig:real-example-acoustic-sensing}
    }
    \caption{\Updated{%
    When an intelligent vehicle approaches a narrow urban intersection,
    \protect\subref{fig:schematic-line-of-sight-sensing}
    traditional line-of-sight sensors cannot detect approaching traffic 
    due to occlusion,
    while \protect\subref{fig:schematic-acoustic-sensing}
    acoustic cues can provide early warnings.
    \protect\subref{fig:real-example-acoustic-sensing} Real-time beamforming
    reveals reflections of the acoustic signal on the walls, especially salient
    on the side opposing the approaching vehicle.
    Learning to recognize these patterns from data enables detection before line-of-sight.
    }}
    \label{fig:schematic}
\end{figure}

\IEEEPARstart{H}{ighly}
automated and self-driving vehicles currently rely on three complementary main sensors to identify visible objects, namely camera, lidar, and radar.
However, the capabilities of these conventional sensors can be limited in urban environments
when sight is obstructed by narrow streets, trees, parked vehicles, and other traffic.
Approaching road users may therefore remain undetected by the main sensors, resulting in dangerous situations and last-moment emergency maneuvers~\cite{keller2011active}.
While future wireless vehicle-to-everything communication (V2X) might mitigate this problem, creating a robust
omnipresent communication layer is still an open problem~\cite{machardy2018v2x}
and excludes road users without wireless capabilities.
Acoustic perception does not rely on line-of-sight
and provides a wide range of complementary and important cues on nearby traffic: 
There are salient sounds with specified meanings, e.g. sirens, car horns,
and reverse driving warning beeps of trucks,
but also inadvertent sounds from tire-road contact and engine use.

In this work,
we propose to use multiple cheap microphones to capture sound as an auxiliary sensing modality
for early detection of approaching vehicles behind blind corners in urban environments.
Crucially, we show that a data-driven pattern recognition approach can successfully identify such situations from the acoustic reflection patterns on building walls
and provide early warnings
before conventional line-of-sight sensing is able to
(see Figure~\ref{fig:schematic}).
While a vehicle should always exit narrow streets
cautiously,
early warnings would reduce the risk of a last-moment emergency brake.

%% file: tex/relatedworks.tex
\Updated{
We here focus on passive acoustic sensing
in mobile robotics~\cite{argentieri2015survey,rascon2017localization,wang2018acoustic}
to detect and localize nearby sounds,
which we distinguish from active acoustic sensing using self-generated sound signals, e.g.~\cite{lindell2019acoustic}.
}
While mobile robotic platforms in outdoor environments may
suffer from vibrations and wind,
various works have demonstrated detection and localization of salient sounds on moving drones~\cite{okutani2012outdoor}
and
wheeled platforms~\cite{an2018reflection,jang2015development}.

Although acoustic cues are known to be crucial for traffic awareness by
pedestrians and cyclist~\cite{stelling2015traffic},
only few works
have explored passive acoustic sensing as a sensor for Intelligent Vehicles (IVs).
\cite{jang2015development,mizumachi2014robust,padmanabhan2014acoustics}
focus on detection and tracking in direct line-of-sight.
\cite{asahi2011development,singh2012non} address detection behind corners from a static observer.
\cite{asahi2011development} only shows experiments without directional estimation.
\cite{singh2012non} tries to accurately model wave refractions,
but experiments in an artificial lab setup show limited success.
Both \cite{asahi2011development,singh2012non}
rely on strong modeling assumptions, ignoring that other informative patterns could be present in the acoustic data.
Acoustic traffic perception is furthermore used for road-side traffic monitoring, e.g. to count vehicles and estimate traffic density
\cite{toyoda2014traffic,ishida2018saved}.
While the increase in Electric Vehicles (EVs) may reduce overall traffic noise,  \cite{sandberg2010vehicles} shows that at 20-30km/h the noise levels for EV and internal combustion vehicles are already similar due to tire-road contact.
\cite{iversen2015measurement} finds that at lower speeds the difference is only about 4-5 dB,
though many EVs also suffer from audible narrow peaks in the spectrum.
As low speed EVs can impact acoustic awareness of humans too~\cite{stelling2015traffic},
legal minimum sound requirements for EVs are being proposed~\cite{robart2013evader,lee2017objective}.

Direction-of-Arrival 
estimation is a key task for sound source localization,
and over the past decades many algorithms have been proposed~\cite{argentieri2015survey,scheibler2018pyroomacoustics}, %
such as the Steered-Response Power Phase Transform (SRP-PHAT)~\cite{dibiase2000high} which is well-suited for reverberant environments with possibly distant unknown sound sources.
Still, in urban settings nearby walls, corners, and surfaces distort sound signals through reflections and
diffraction~\cite{hornikx2011modelling}.
Accounting for such distortions has shown to improve localization~\cite{an2018reflection,zhang2017surround},
but only in controlled indoor environments where \Updated{detailed} knowledge of the surrounding geometry is available.

Recently, data-driven methods have shown promising results in challenging real-world conditions for
various acoustic tasks.
For instance, learned sound models assist monaural source separation~\cite{osako2017supervised} and source localization from direction-dependent attenuations by fixed structures~\cite{saxena2009learning}. 
Increasingly, deep learning is used for audio classification~\cite{salamon2017deep,valada2018deep},
\Updated{and localization~\cite{yalta2017}
of sources in line-of-sight, in which case visual detectors can replace manual labeling~\cite{he2018deep, gan19}}.
Analogous to our work, \cite{scheiner2020seeing}
presents a first deep learning method for sensing around corners but with automotive radar.
Thus, while the effect of occlusions on sensor measurements is difficult to model \cite{singh2012non},
data-driven approaches appear to be a good alternative.
\\

This paper provides the following contributions:
First,
we demonstrate in real-world outdoor conditions that a vehicle-mounted microphone array can detect \Updated{the sound of approaching vehicles behind blind corners
from reflections on nearby surfaces 
before line-of-sight detection is feasible.}
This is a key advantage for IVs,
where passive acoustic sensing is still relatively under-explored.
Our experiments investigate the impact on accuracy and detection time for various conditions, such as different acoustic environments, driving versus static ego-vehicle,
and compare to current visual and acoustic baselines.

Second, we propose a data-driven detection pipeline to efficiently address this task
and show that it outperforms model-driven acoustic signal processing.
\Updated{
Unlike existing data-driven approaches, we cannot use visual detectors for positional labeling~\cite{he2018deep} or transfer learning~\cite{gan19}, since our targets are visually occluded.
Instead, we cast the task as a multi-class classification problem
to identify if and from what corner a vehicle is approaching.
We demonstrate that Direction-of-Arrival estimation can provide robust
features to classify sound reflection patterns,
even without end-to-end feature learning and large amounts of data.}

Third,
for our experiments we collected a new audio-visual dataset in real-world urban environments.\footnote{%
\Updated{Code \& data: \href{https://github.com/tudelft-iv/occluded_vehicle_acoustic_detection}{github.com/tudelft-iv/occluded\_vehicle\_acoustic\_detection}}}
To collect data, we mounted a front-facing microphone array on our research vehicle, which additionally has a front-facing camera. This prototype setup facilitates qualitative and quantitative experimentation of different acoustic perception tasks.

%% file: tex/methods.tex
\label{sec:approach}

Ideally, an ego-vehicle driving through an area with occluding structures is able to early predict \textit{if} and from \textit{where} another vehicle is approaching, even if it is from behind a blind corner as illustrated in Figure~\ref{fig:schematic}.
Concretely, this work aims to distinguish three situations as early as possible using ego-vehicle sensors only:
\begin{itemize}
    \item an occluded vehicle approaches from behind a corner on the \textit{left}, and only moves into view last-moment when the ego-vehicle is about to reach the junction,
    \item same, but vehicle approaches behind a \textit{right} corner,
    \item no vehicle is approaching.
\end{itemize}

We propose to consider this task an online classification problem.
As the ego-vehicle approaches a blind corner, the acoustic measurements made over short time spans should be assigned to one in a set of four classes,
$\classset{} = $ \{\classleft{}, \classfront{}, \classright{}, \classneg{}\},
where \classleft{}/\classright{} indicates a still occluded (i.e. not yet in direct line-of-sight) approaching vehicle behind a corner on the left/right,
\classfront{} that the vehicle is already in direct line-of-sight,
and \classneg{} that no vehicle is approaching.

In Section~\ref{sec:baselines} we shall first consider two line-of-sight baseline approaches for  detecting vehicles.
Section~\ref{sec:occ-vehicle-det} then elaborates our proposed extension to acoustic non-line-of-sight detection.
Section~\ref{sec:hardware} provides details of our vehicle's novel acoustic sensor setup used for data collection.

\begin{figure*}[h]
    \centering
	     \includegraphics[width=.95\textwidth]{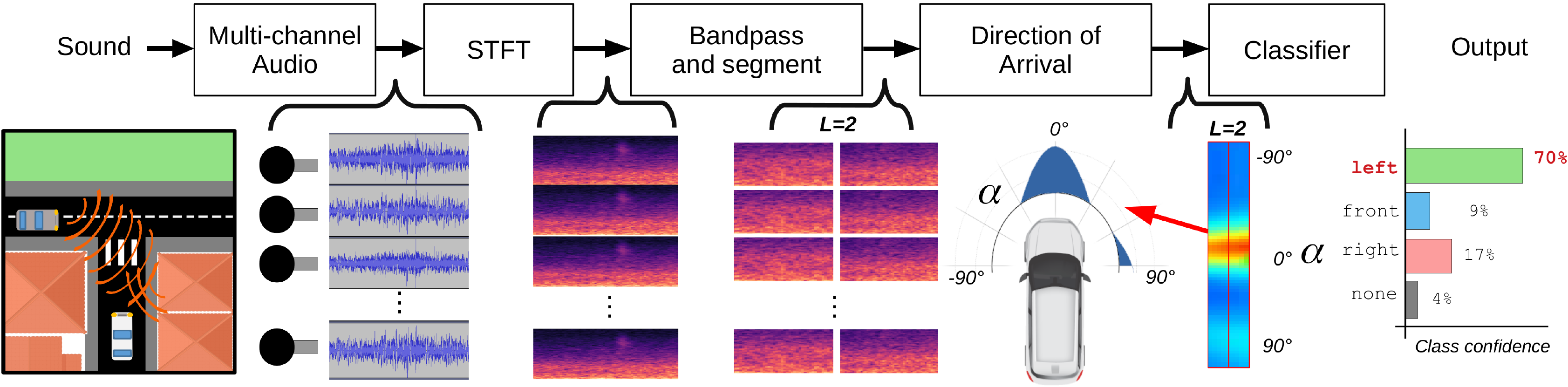}
    \caption{Overview of our acoustic detection pipeline, see Section~\ref{sec:occ-vehicle-det} for an explanation of the steps.}
    \label{fig:detection-flowchart}
\end{figure*}

\subsection{Line-of-sight detection}
\label{sec:baselines}

We first consider how the task would be addressed with line-of-sight vehicle detection using either conventional cameras, or using past work on acoustic vehicle detection. 

\paragraph{Visual detection baseline}
Cameras
are currently one of the de-facto choices
for detecting
vehicles and other objects within line-of-sight.
Data-driven Convolutional Neural Networks have proven to be highly effective on images.
However, visual detection can  only detect vehicles that are already (partially) visible, and thus only distinguishes between \classfront{} and \classneg{}.
To demonstrate this,
we use
Faster R-CNN~\cite{ren2015faster},
a state-of-the-art visual object detector,
on the ego-vehicle's front-facing camera as a visual baseline.

\paragraph{Acoustic detection baseline}
Next, we consider that the ego-vehicle is equipped with an array of $\nummics{}$ microphones.
\Updated{As limited training data hinders learning features (unlike~\cite{he2018deep,gan19}),}
we leverage beamforming to estimate the Direction-of-Arrival (DoA)
of tire and engine sounds originating from the approaching vehicle.
DoA estimation directly identifies the presence and direction of such sound sources,
and has been shown to work robustly in unoccluded conditions \cite{mizumachi2014robust,jang2015development}.
Since sounds can be heard around corners,
and low frequencies diffract (``bend'') around corners~\cite{hornikx2011modelling},
one might wonder: Does the DoA of the sound of an occluded vehicle correctly identify from where the vehicle is approaching?
To test this hypothesis for our target real-world application,
our second baseline follows~\cite{mizumachi2014robust,jang2015development} and directly uses the most salient DoA angle estimate.

Specifically, the implementation uses the Steered-Response Power-Phase Transform (SRP-PHAT) \cite{dibiase2000high} for DoA estimation.
SRP-PHAT
relates the spatial layout of sets of microphone pairs and the temporal offsets of
the corresponding audio signals to their relative distance to the sound source.
To apply SRP-PHAT on $\nummics{}$ continuous synchronized signals,
only the most recent $\samplelen{}$ seconds are processed.
On each signal, a Short-Time Fourier Transform (STFT)
is computed with a Hann windowing function,
and a frequency bandpass for the $[\minfreq{}, \maxfreq{}]$ Hz range.
Using the generalized cross-correlation of the $\nummics{}$ STFTs,
SRP-PHAT computes the DoA energy $r(\alpha)$ for any given azimuth
angle $\alpha$ around the vehicle. Here $\alpha = -90^\circ/0^\circ/+90^\circ$ indicates an angle towards the left/front/right of the vehicle respectively.
If the hypothesis holds that the overall salient sound direction $\alpha_{\mathrm{max}} = \argmax r(\alpha)$ remains intact due to diffraction,
one only needs to determine if $\alpha_{\mathrm{max}}$ is beyond some sufficient threshold  $\alphathreshold{}$.
The baseline thus assigns class \classleft{} if $\alpha_{\mathrm{max}} < -\alphathreshold{}$,
\classfront{} if $-\alphathreshold{} \leq \alpha_{\mathrm{max}} \leq +\alphathreshold{}$, and \classright{} if $\alpha_{\mathrm{max}} > +\alphathreshold{}$.
We shall evaluate this baseline on the easier task of only separating these three classes, and ignore the \classneg{} class.

\subsection{Non-line-of-sight acoustic detection}
\label{sec:occ-vehicle-det}

We argue that in contrast to line-of-sight detection,
DoA estimation alone is unsuited for occluded vehicle detection %
(and confirm this in Section~\ref{sec:classifier}).
Salient sounds produce sound wave reflections on surfaces, such as walls (see Figure \ref{fig:real-example-acoustic-sensing}), thus the DoA does not indicate the actual location of the source.
\Updated{%
Modelling the sound propagation~\cite{an2018reflection} while driving through uncontrolled outdoor environments is challenging, especially as accurate models of the local geometry are missing.
Therefore, we take a data-driven approach and treat the \textit{full energy distribution}
from SRP-PHAT as robust features for our classifier that capture all reflections.
}

An overview of the proposed processing pipeline is shown in Figure~\ref{fig:detection-flowchart}.
We again create $\nummics{}$ STFTs,
using a temporal windows of $\samplelen{}$ seconds,
Hann windowing function
and a frequency bandpass of $[\minfreq{}, \maxfreq{}]$ Hz.
Notably, we do not apply any other form of noise filtering or suppression.
To capture temporal changes in the reflection pattern, 
we split the STFTs along the temporal dimension
into $\numsegments{}$ non-overlapping segments.
For each segment, we compute the DoA energy
at multiple azimuth angles $\alpha$ in front of the vehicle.
The azimuth range $[-90^\circ, +90^\circ]$ is divided into $B$ equal bins $\alpha_{1}, \cdots, \alpha_{B}$.
From the original $M$ signals,
we thus obtain $\numsegments{}$ response vectors $\boldsymbol{r}_l = [r_l(\alpha_1), \cdots, r_l(\alpha_B)]^\top$.
Finally, these are concatenated to a $(\numsegments{} \times B)$-dimensional feature vector $\boldsymbol{x} = [\boldsymbol{r}_{1}, \cdots, \boldsymbol{r}_{L}]^\top$,
for which a Support Vector Machine is trained to predict $\classset{}$.
Note that increasing the temporal resolution by having more segments $\numsegments{}$
comes at the trade-off of a increased final feature vector size and reduced DoA estimation quality due to shorter time windows.

\subsection{Acoustic perception research vehicle}
\label{sec:hardware}

\begin{figure}[ht]
    \centering
    \scalebox{0.9}{
        \begin{tikzpicture}
    	\node[anchor=south west,inner sep=0] (image) at (0,0) 	{\includegraphics[width=.47\textwidth]{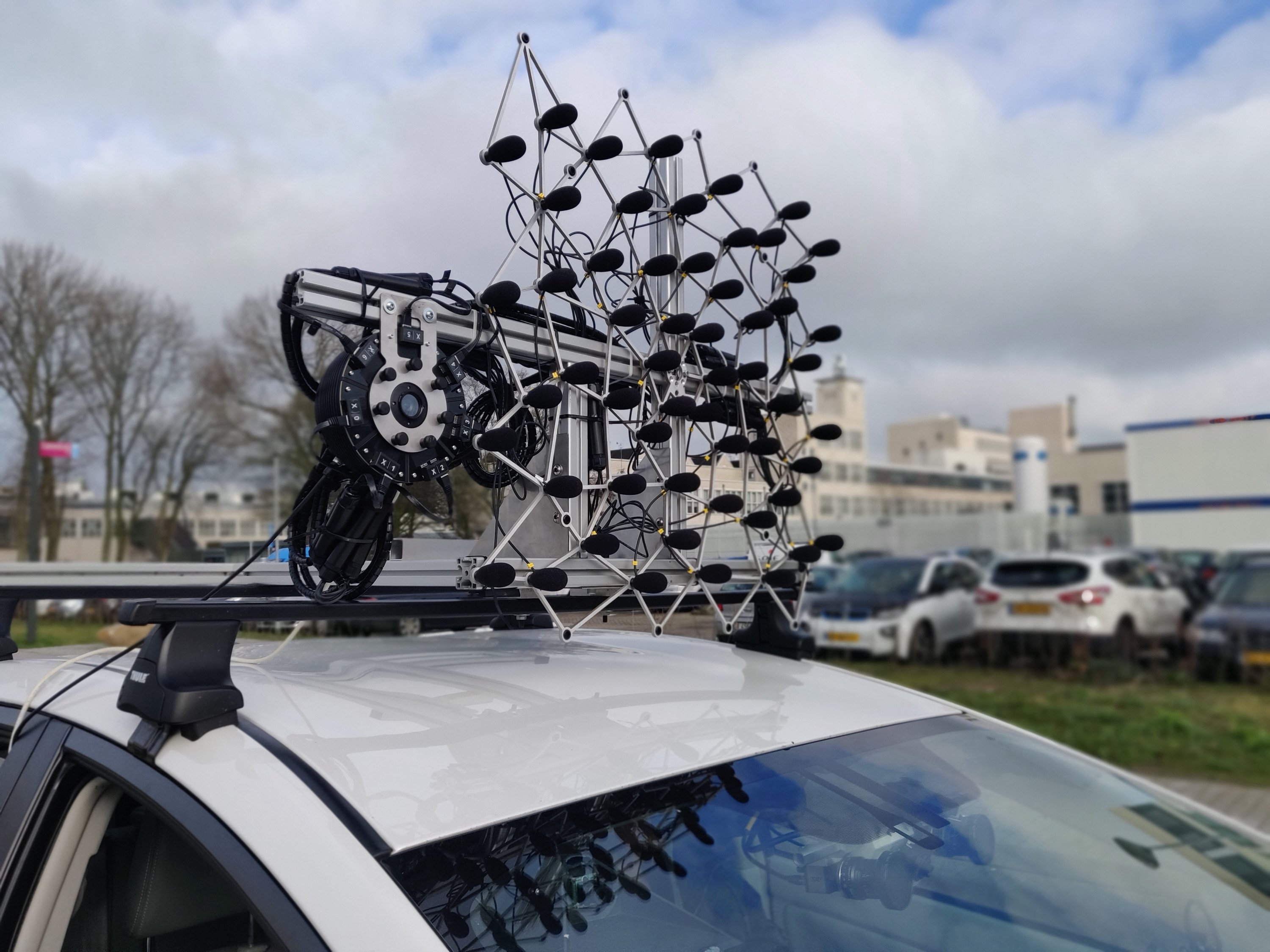}};
    	\begin{scope}[x={(image.south east)},y={(image.north west)}]
    	\node[white,fill=black!100] at (0.80,0.80) (a){\huge \bf A};
    	\draw [-latex, ultra thick, red] (a) to (0.55,0.6);
    	\node[white,fill=black!100] at (0.10,0.58) (b){\huge \bf B};
    	\draw [-latex, ultra thick, red] (b) to (0.25,0.58);
    	\node[white,fill=black!100] at (0.5,0.25) (c){\huge \bf C};
    	\draw [-latex, ultra thick, red] (c) to (0.7,0.13);
    	\end{scope}
    	\node[anchor=south west,inner sep=0] (image) at (0,4.6) 
    	{\includegraphics[width=.1\textwidth]{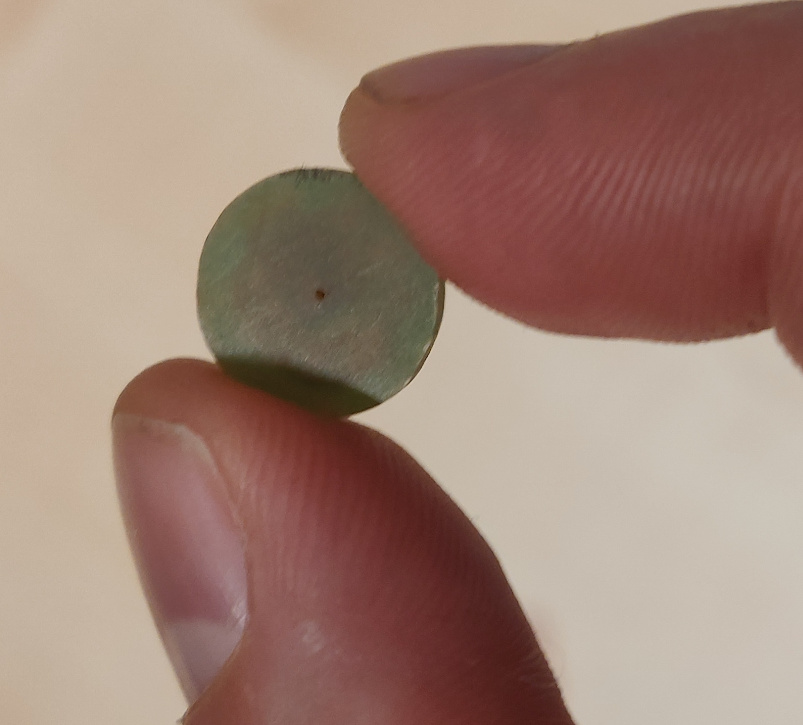}};
    	\end{tikzpicture}
	}
    \caption{Sensor setup of our test vehicle. A: Center of the 56 MEMS acoustic array. B: signal processing unit. C: front camera behind windscreen.
    Inset: the diameter of a single MEMS microphone is only 12mm.}
    \label{fig:prius-with-micarray}	
\end{figure}
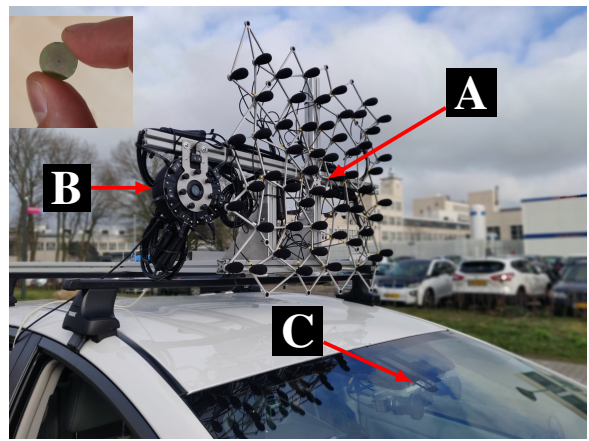
\begin{figure*}[h]
    \centering
    \subfloat[Stroller at a distance]
    {\includegraphics[width=0.18\linewidth] {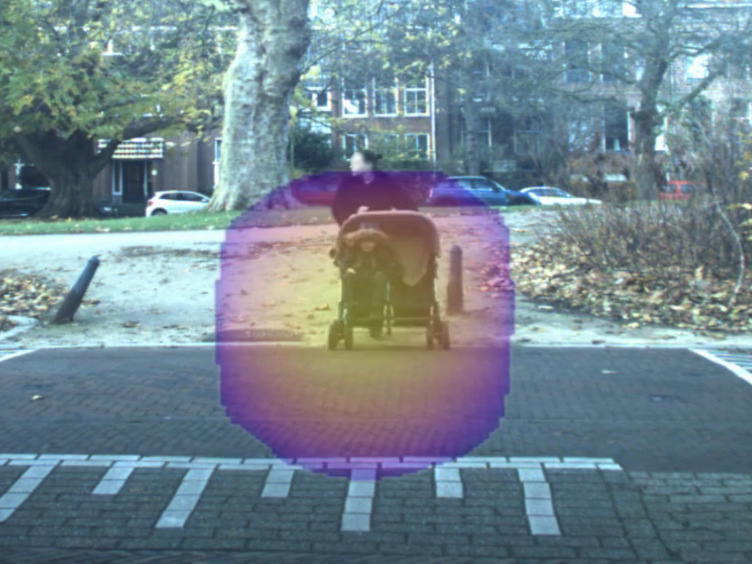}
    \label{fig:beamforming-example1}
    }
    \subfloat[Electric scooter]
    {\includegraphics[width=0.18\linewidth] {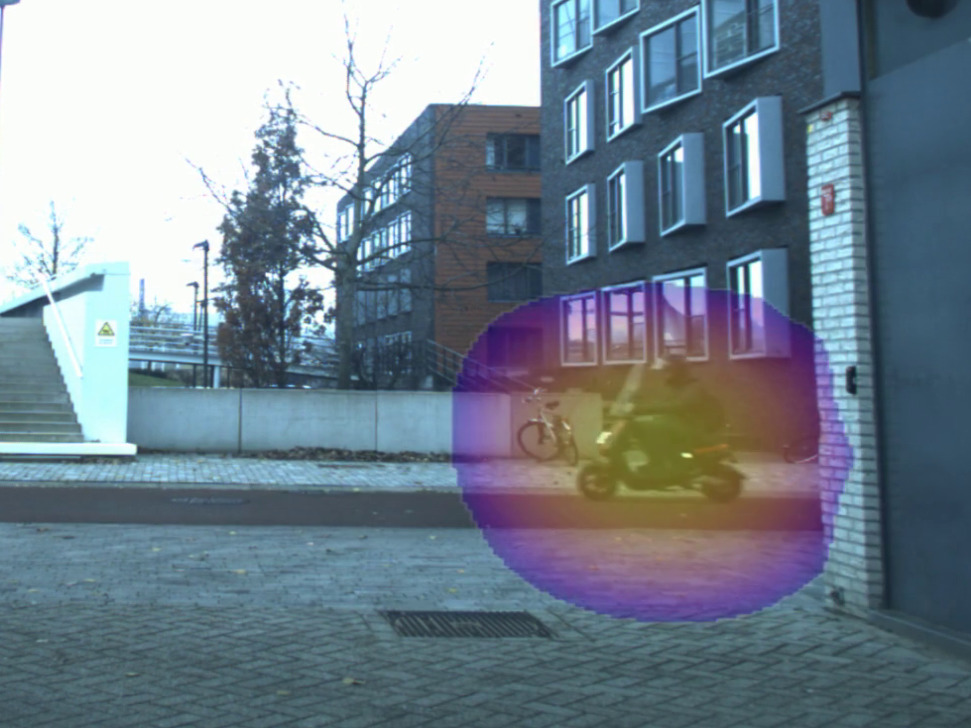}
    \label{fig:beamforming-example2}
    }
    \subfloat[Scooter overtaking]
    {\includegraphics[width=0.18\linewidth] {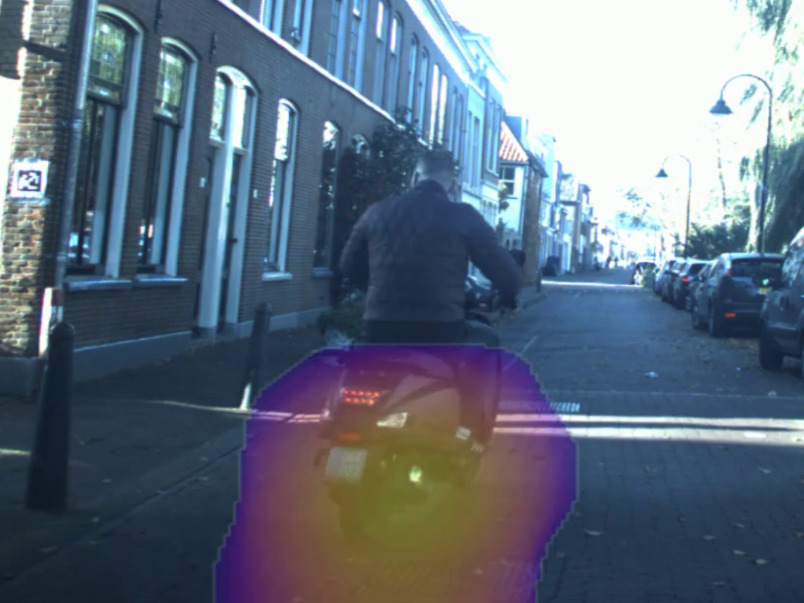}    \label{fig:beamforming-example3}
    }
    \subfloat[Car passing by]
    {\includegraphics[width=0.18\linewidth] {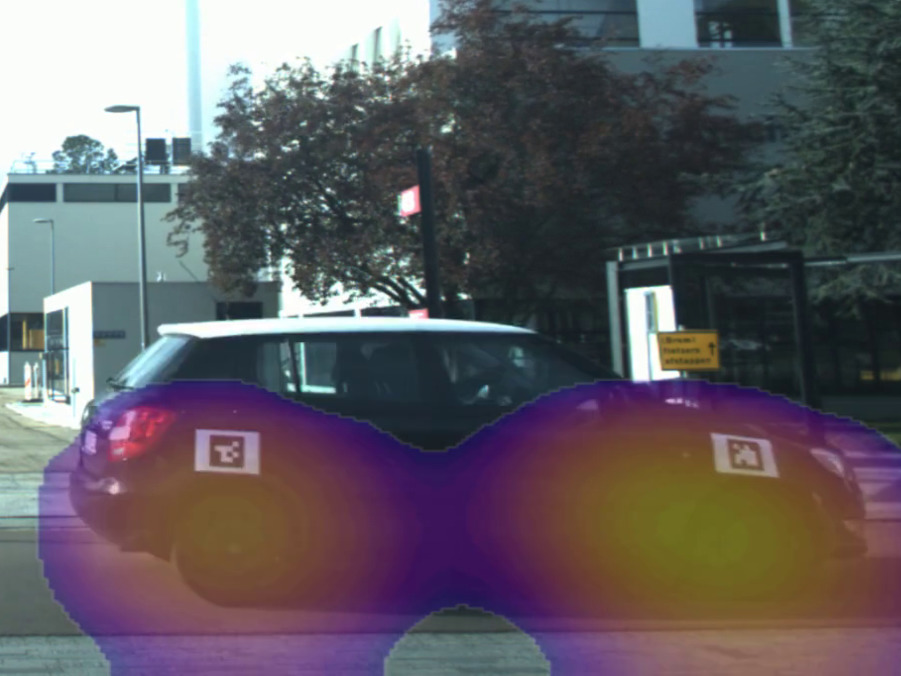}
    \label{fig:beamforming-example4}
    }
    \subfloat[Oncoming car]
    {\includegraphics[width=0.18\linewidth] {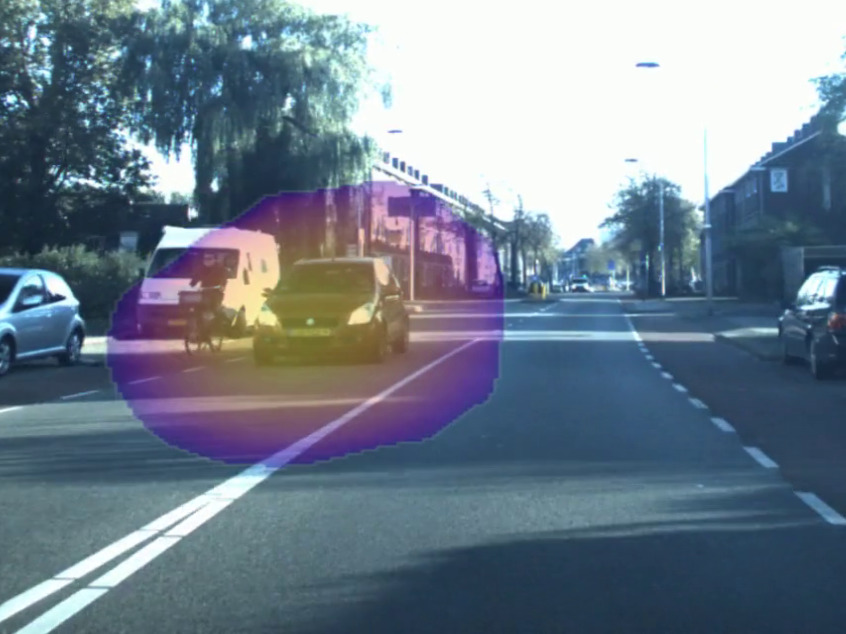}
    \label{fig:beamforming-example5}
    }
    \caption{Qualitative examples of 2D Direction-of-Arrival estimation overlaid on the camera image (zoomed).
    \protect\subref{fig:beamforming-example1}: Stroller wheels are picked up even at a distance.
    \protect\subref{fig:beamforming-example2},
    \protect\subref{fig:beamforming-example3}:
    Both conventional and more quiet electric scooters are detected.
    \protect\subref{fig:beamforming-example4}:
    The loudest sound of a passing vehicle is typically the road contact of the individual tires.
    \protect\subref{fig:beamforming-example5}: 
    Even when the ego-vehicle drives at $\sim 30$ km/h, oncoming moving vehicles are still registered as salient sound sources.
    }%
    \label{fig:beamforming-examples}
\end{figure*}

To collect real-world data and demonstrate non-line-of-sight detection,
a custom microphone array was mounted
on the roof rack of our research vehicle~\cite{ferranti2019safevru},
a hybrid electric Toyota Prius.
The microphone array hardware consists of 56 ADMP441 MEMS microphones, supports data acquisition at 48 kHz sample rate, 24 bits resolution, and synchronous sampling.
It was bought from \textit{CAE Software \& Systems GmbH} with a metal frame.
On this $0.8m \times 0.7m$ frame the microphones are distributed semi-randomly 
while the microphone density remains homogeneous.
The general purpose layout was designed by the company through stochastic optimization to have large variance in inter-microphone distances
and serve a wide range of acoustic imaging tasks.
The vehicle is also equipped with a front-facing camera for data collection and processing.
The center of the microphone array is about 1.78m above the ground, and 0.54m above and 0.50m behind the used front camera, see Figure~\ref{fig:prius-with-micarray}.
\Updated{As depicted in the Figure's inset, the microphones themselves are only 12mm wide.
They cost about US\$1 each.}

\Updated{A signal processing unit receives the analog microphone signals, and sends the data over Ethernet to a PC running the Robot Operating System (ROS).
Using ROS, the synchronized microphone signals are collected together with other vehicle sensor data.} %
Processing is done in python, using \textit{pyroomacoustics}~\cite{scheibler2018pyroomacoustics}
for acoustic feature extraction,
and \textit{scikit-learn}~\cite{pedregosa2011scikit} for classifier training.

\Updated{We emphasize that this setup is not intended as a production prototype,
but provides research benefits:
The 2D planar arrangement provides both horizontal and vertical high-resolution
DoA responses, which can be overlaid as 2D heatmaps~\cite{sarradj2017acoular} on the front camera image
to visually study the salient sources (Section~\ref{sec:qualitative-results}).
By testing subsets of microphones, we can assess the impact of the number of microphones and their relative placement (Section~\ref{sec:number-of-mics}).
In the future, the array should only use a few microphones at various locations around the vehicle.}

%% file: tex/experiments.tex
To validate our method, we created a novel dataset with our acoustic research vehicle in real-world urban environments.
We first illustrate the quality of acoustic beamforming in such conditions
before turning to our main experiments.

\subsection{Line-of-sight localization -- qualitative results}
\label{sec:qualitative-results}
As explained in Section \ref{sec:hardware}, the heatmaps of the 2D DoA results
can be overlaid with the camera images.
Figure~\ref{fig:beamforming-examples}
shows some interesting qualitative findings in real urban conditions.
The examples
highlight that beamforming can indeed pick up various important acoustic events for autonomous driving
in line-of-sight, such as the presence of 
vehicles and some vulnerable road users (e.g. strollers).
Remarkably, even electric scooters and oncoming traffic \textit{while the ego-vehicle is driving}
are recognized as salient sound sources.
\Updated{A key observation from Figure~\ref{fig:real-example-acoustic-sensing} is that sounds originating behind corners reflect in particular patterns on nearby walls.}
Overall, these results show the feasibility of acoustic detection of (occluded) traffic.

\subsection{Non-line-of-sight dataset and evaluation metrics}
\label{sec:dataset}

\begin{figure}[b]
\centering
\subfloat[\textbf{Type A}: completely walled]
  {\includegraphics[width=0.2\textwidth]{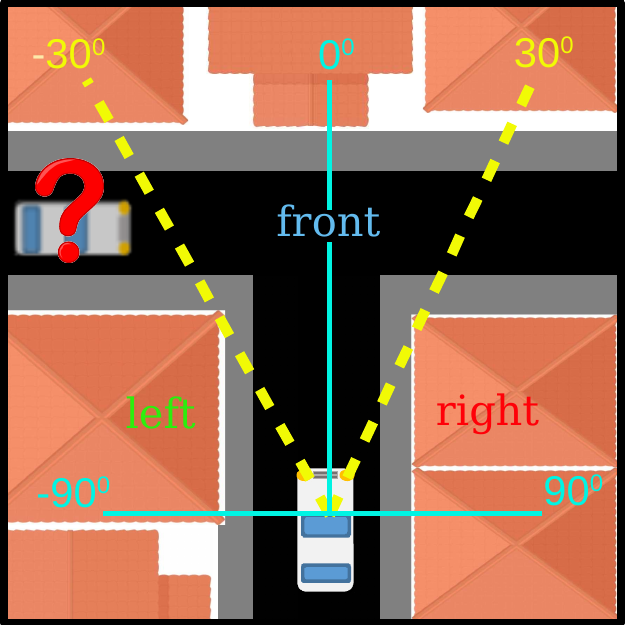}}~
\subfloat[\textbf{Type B}: walled exit]
  {\includegraphics[width=0.2\textwidth]{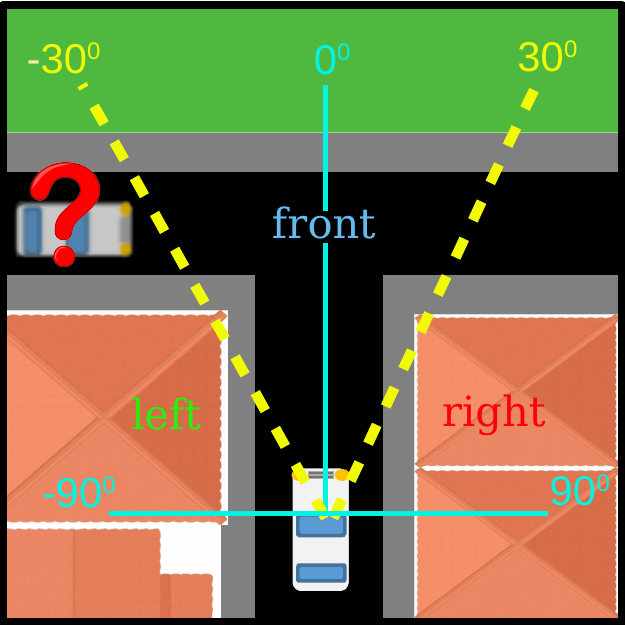}}
\caption{Schematics of considered environment types.
The ego-vehicle approaches the junction from the bottom.
Another vehicle might approach behind the left or right blind corner.
Dashed lines indicate the camera FoV.
}
\label{fig:locationtypes}
\end{figure}

The quantitative experiments are designed to separately control and study various
factors that could influence acoustic perception.
We collected multiple recordings of the
situations explained in Section \ref{sec:approach}
at five T-junction locations with blind corners in the inner city of Delft.
The locations are categorized into two types of walled acoustical environments, namely types A and B (see Figure~\ref{fig:locationtypes}).
At these locations common background noise, such as construction sites and other traffic,
was present at various volumes.
For safety and control, we did not record in the presence of other motorized traffic on the roads at the target junction.

The recordings can further be divided into \textit{Static} data, made while is the ego-vehicle in front of the junction but not moving,
and more challenging \textit{Dynamic} data where the ego-vehicle reaches the junction at $\sim$15 km/h (see the supplementary video).
Static data is easily collected,
and ensures that the main source of variance is the approaching vehicle's changing position.

For the static case, 
the ego-vehicle was positioned such that the building corners are still visible in the camera and
occlude the view onto the intersecting road (on average a distance of $\sim$7-10m from the intersection).
Different types of passing vehicles were recorded,
although in most recordings the approaching vehicle was a \v{S}koda Fabia 1.2 TSI (2010) driven by one of the authors.
For the Dynamic case, coordinated recordings with the \v{S}koda Fabia were conducted
to ensure that encounters were relevant and executed in a safe manner.
Situations with \classleft{}/\classright{}/\classneg{} approaching vehicles were performed in arbitrary order to prevent undesirable \Updated{correlation of background noise to some class labels}.
In $\sim$70\% of the total Dynamic recordings and $\sim$19.5\% of the total Static recordings, the ego-vehicle's noisy internal combustion engine was running to charge its battery.
\begin{table}[h]
    \caption{Samples per subset. In the ID, S/D indicates Static/Dynamic ego-vehicle, A/B the environment type (see figure~\ref{fig:locationtypes}).}
    \label{tab:C_dataset}
        \begin{tabular}{l|cccc|c}
        ID & \classleft{} & \classfront{} & \classright{} & \classneg{} & Sum \\ %
        \hline
        SA1 / DA1 & 14 / 19  & 30 / 38 & 16 / 19 & 30 / 37 & ~90/113 \\ %
        SA2 / DA2 & 22 / ~7  & 41 / 15 & 19 / ~8 & 49 / 13 & 131/~43 \\
        SB1 / DB1 & 17 / 18 & 41 / 36 & 24 / 18 & 32 / 35 & 114/107 \\ %
        SB2 / DB2 & 28 / 10 & 55 / 22 & 27 / 12 & 43 / 22 & 153/~66 \\ %
        SB3 / DB3 & 22 / 19 & 45 / 38 & 23 / 19 & 45 / 36 & 135/112 \\ %
        \hline \hline
        SAB / DAB & 103/~73 & 212/149 & 109/~76 & 199/143 & \textit{\nostaticrecordings{}}/\textit{\nodrivingrecordings{}} \\ %
        \end{tabular}
\end{table}

\paragraph{Sample extraction}
For each Static recording with an approaching target vehicle,
the time $t_0$ is manually annotated as the moment when the approaching vehicle enters direct line-of-sight.
Since the quality of our $t_0$ estimate is bounded by the ego-vehicle's camera frame rate (10 Hz),
we conservatively regard the last image \textit{before} the incoming vehicle is visible as $t_0$.
Thus, there is no line-of-sight at $t \le t_0$. 
At $t > t_0$ the vehicle is considered visible, even though it might only be a fraction of the body.
For the Dynamic data, this annotation is not feasible as the approaching car may be in direct line-of-sight,
yet outside the limited field-of-view of the front-facing camera as the ego-vehicle has advanced onto the intersection.
Thus, annotating $t_0$ based on the camera images is not representative for line-of-sight detection.
To still compare our results across locations, we manually annotate the time $\tau_0$,
the moment when the ego-vehicle is at the same position as in the corresponding Static recordings.
All Dynamic recordings are aligned to that time
as it represents the moment where the ego-vehicle should make a classification decision,
irrespective if an approaching vehicle is about to enter line-of-sight or still further away.

From the recordings, \Updated{short $\samplelen{} = 1\mathrm{s}$ audio samples are extracted.
Let $t_e$, the end of the time window $[t_e-1\mathrm{s}, t_e]$}, denote a sample's
time stamp at which a prediction could be made.
For Static \classleft{} and \classright{} recordings,
samples with the corresponding class label are extracted at $t_e = t_0$.
\Updated{%
For Dynamic recordings,
\classleft{} and \classright{} samples are extracted at $t_e = \tau_0 + 0.5\mathrm{s}$.
This ensures that during the $1\mathrm{s}$ window the ego-vehicle is on average close to its position in the Static recordings.
In both types of recordings, \classfront{} samples are extracted 
1.5s after the \classleft{}/\classright{} samples, e.g.~$t_e = t_0 + 1.5 \mathrm{s}$.
Class \classneg{} samples were from recordings with no approaching vehicles.
Table \ref{tab:C_dataset} lists statistics of the extracted samples at each recording location}.

\paragraph{Data augmentation}
Table~\ref{tab:C_dataset} shows that the data acquisition scheme produced imbalanced class ratios,
with about half the samples for \classleft{}, \classright{} compared to \classfront{}, \classneg{}.
Our experiments therefore explore \textit{data augmentation}.
By exploiting the symmetry of the angular DoA bins,
augmentation will double the \classright{} and \classleft{} class samples by reversing the azimuth bin order in all $\boldsymbol{r}_l$, resulting in new features for the opposite label,
i.e.~as if additional data was collected at mirrored locations.
\Updated{Augmentation is a training strategy only, and thus not applied to test data to keep results comparable, and distinct for \classleft{} and \classright{}.}

\paragraph{Metrics}
We report the overall accuracy,
and the per-class Jaccard index (a.k.a.~Intersection-over-Union)
as a robust measure of one-vs-all performance.
First, for each class $c$ the True Positives/Negatives ($TP_c$/$TN_c$), and False Positives/Negatives ($FP_c$/$FN_c$)
are computed, treating target class $c$ as positive and the other three classes jointly as negative.
Given the total number of test samples $\numtestdata{}$,
the overall accuracy is then
$\left( \sum_{c \in \classset{}} TP_c \right) / \numtestdata{}$
and the per-class Jaccard index is
$J_c = TP_c / (TP_c + FP_c + FN_c)$.

 \begin{table}[h]
     \caption{Baseline comparison and hyperparameter study w.r.t. our reference configuration: SVM $\lambda=1$, $\samplelen{} = 1$, $\numsegments{}=2$, data augmentation.
     Results on Static data.
     * denotes \textit{our} pipeline.
     }
     \label{tab:eval-all-ablation}
     \label{tab:eval-all-classifier}
 \begin{tabular}{c|c|cccc}
         Run & Accuracy & $J_{\classleft{}}$ & $J_{\classfront{}}$ & $J_{\classright{}}$ & $J_{\classneg{}}$ \\
         \hline
         * \textit{(reference)}                  & \textbf{0.92} & 0.79 & 0.89 & \textbf{0.87} & 0.83 \\
         * wo. data augment.                      & \textbf{0.92} & 0.75 & 0.91 & 0.78 & 0.83 \\
         * w. $\samplelen{} = 0.5 \textrm{s}$ & 0.91 & 0.75 & 0.89 & \textbf{0.87} & 0.82 \\
         * w. $\numsegments{}=1$                   & 0.86 & 0.64 & 0.87 & 0.73 & 0.79 \\
         * w. $\numsegments{}=3$                   & \textbf{0.92} & 0.74 & 0.92 & 0.82 & 0.81 \\
         * w. $\numsegments{}=4$                   & 0.90 & 0.72 & 0.90 & 0.77 & 0.83 \\
         * w. SVM $\lambda=0.1$         & 0.91 & 0.78 & 0.89 & 0.81 & 0.82 \\
         * w. SVM $\lambda=10$          & 0.91 & \textbf{0.81} & 0.86 & 0.84 & 0.83  \\
         \hline
         DoA-only~\cite{mizumachi2014robust,jang2015development} &
         0.64 & 0.11 & 0.83 & 0.28 & - \\ %
         Faster R-CNN~\cite{detectron2} & 0.60 & 0.00 & \textbf{0.99} & 0.00 & \textbf{0.98}
         \end{tabular}
 \end{table}

\subsection{Training and impact of classifier and features}
\label{sec:classifier}

First, the overall system performance and hyperparameters are evaluated on all Static data from both type A and B locations (i.e.~subset ID `SAB') using 5-fold cross-validation.
The folds are fixed once for all experiments,
with the training samples of each class equally distributed among folds.

We fix the frequency range to $\minfreq{} = 50\mathrm{Hz}, \maxfreq{} = 1500\mathrm{Hz}$,
and the number of azimuth bins to $B = 30$ (Section \ref{sec:occ-vehicle-det}).
For efficiency and robustness,
a linear Support Vector Machine (SVM) is used
with $l2-$regularization weighted by hyperparameter $\lambda$.
Other hyperparameters to explore include the sample length $\samplelen{} \in \{0.5\mathrm{s}, 1\mathrm{s}\}$,
the segment count $\numsegments{} \in \{1,2,3,4\}$,
and using/not using data augmentation.

Our final choice and reference is the SVM with $\lambda = 1$,
$\samplelen{}=1\mathrm{s}$, $\numsegments{} = 2$, and data augmentation.
Table \ref{tab:eval-all-ablation} shows the results for changing these parameter choices.
\Updated{The overall accuracy for all these hyperparameters choices is mostly similar,
though per-class performance does differ.
Our reference achieves top accuracy,
while also performing well on both \classleft{} and \classright{}.
We keep its hyperparameters for all following experiments.}

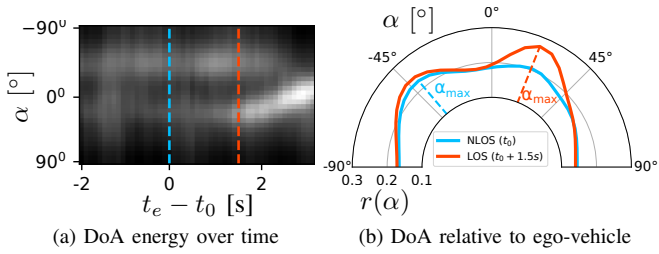
\begin{figure}[t]
\centering
\subfloat[DoA energy over time]
{\begin{tikzpicture}
      \node[anchor=south west, inner sep=0] (image1) at (0,0)
      {\includegraphics[width=0.44\linewidth]{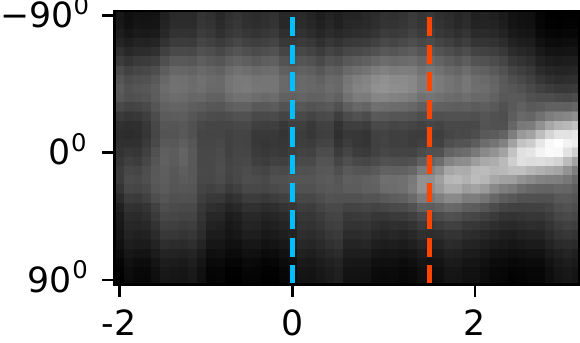}};
      \node[anchor=south,below=of image1,inner sep=0,xshift=0.4cm, yshift=1.0cm] 
    	        {$t_e-t_0$ [s]}; %
     \node[inner sep=0,rotate=90,left=of image1,yshift=-1.0cm, xshift=0.5cm]
    	        {$\alpha~[{}^\circ]$};
  \end{tikzpicture}
}
\subfloat[DoA relative to ego-vehicle]
{\begin{tikzpicture}
    \node[anchor=south west, inner sep=0] (image) at (0,0)
    {\includegraphics[width=0.5\linewidth]{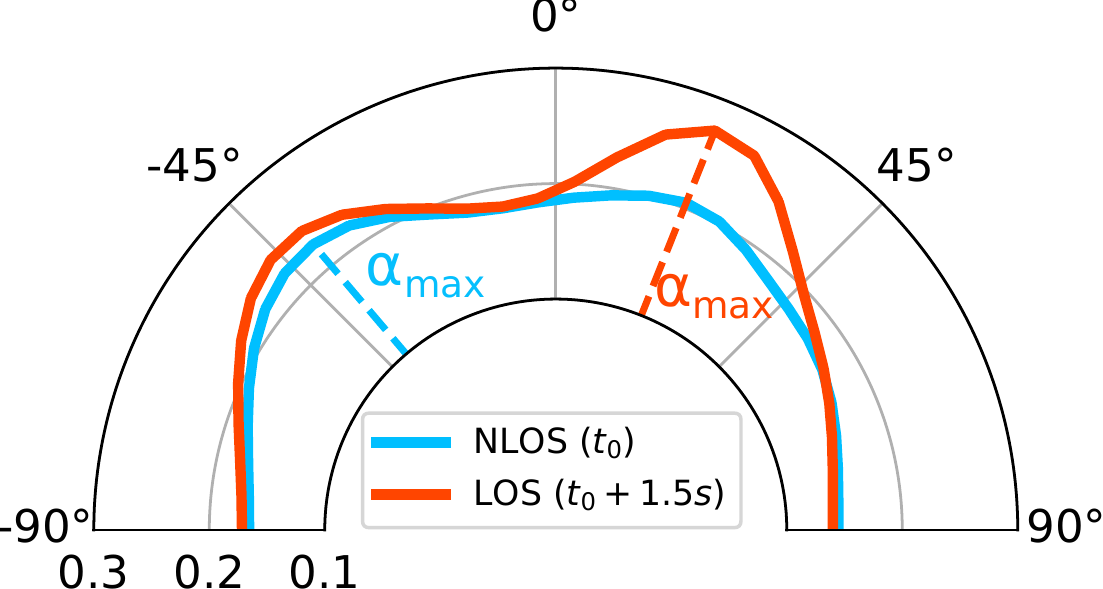}};
    \node[anchor=south,below=of image,inner sep=0, yshift=0.95cm, xshift=-1.4cm]{$r(\alpha)$};
    \node[inner sep=0,rotate=0, left=of image, yshift=1.0cm, xshift=2.5cm ]{$\alpha~[{}^\circ]$};
  \end{tikzpicture}
 }
\caption{\Updated{%
DoA energy over time for the recording shown in Figure \ref{fig:real-example-acoustic-sensing}.
When the approaching vehicle is not in line-of-sight (NLOS), e.g.~at $t_0$, the main peak is a reflection on the
wall ($\alpha_{max} < -30^\circ$) opposite of that vehicle.
}}
\label{fig:doa-over-time}
\end{figure}

The table also shows the results of the DoA-only baseline explained in
Section \ref{sec:baselines} using $\alphathreshold = 50^{\circ}$,
which was found through a grid search in the range $[0^\circ, 90^\circ]$.
As expected, the DoA-only baseline~\cite{mizumachi2014robust,jang2015development} shows weak performance for all metrics.
\Updated{%
While the sound source is occluded, the most salient sound direction does not represent its origin,
but its reflection on the opposite wall (see Figure \ref{fig:schematic}).
The temporal evolution of the full DoA energy for a car approaching from the \classright{} is shown in Figure~\ref{fig:doa-over-time}.
}
When it is still occluded at $t_0$, there are multiple peaks and the most salient one is a reflection on the left ($\alpha_{max} \approx -40^\circ$).
Only once the car is in line-of-sight ($t_0 + 1.5\mathrm{s}$)
the main mode clearly represents its true direction ($\alpha_{max} \approx +25^\circ$).
\Updated{%
The left and right image in Figure \ref{fig:real-example-acoustic-sensing}
also show such peaks at $t_0$ and $t_0 + 1.5\mathrm{s}$, respectively.
}

The bottom row of the table shows the visual baseline,
a Faster R-CNN R50-C4 model
trained on the COCO dataset \cite{detectron2}.
To avoid false positive detections, we set the score threshold of 75\% and additionally
required a bounding box height of 100 pixels to ignore cars far away in the background,
which were not of interest.
Generally this threshold
is already exceeded once the hood of the approaching car is visible. 
While performing well on \classfront{} and \classneg{}, this visual baseline shows poor overall accuracy as
it is physically incapable of classifying \classleft{} and \classright{}.

\subsection{Detection time before appearance}
\label{sec:detection-time}

Ultimately, the goal is to know whether our acoustic method can detect approaching vehicles earlier than the state-of-the-art visual baseline.
For this purpose, their online performance is compared next.

The static recordings are divided into a fixed training (328 recordings) and test (83 recordings) split,
stratified to adequately represent labels and locations.
The training was conducted as in Section \ref{sec:classifier} with 
\classleft{} and \classright{} samples extracted at $t_e = t_0$.
The visual baseline is evaluated on every camera frame (10 Hz).
Our detector is evaluated on a sliding window of $1s$
across the 83 test recordings.
To account for the transition period when the car may still be partly occluded,
\classfront{} predictions by both methods are accepted as correct starting at $t = t_0$.
For recordings of classes \classleft{} and \classright{}, these classes are accepted until $t = t_0+1.5$s, allowing for temporal overlap with \classfront{}.

\begin{figure}[h]
     \begin{tikzpicture}
	     \node[anchor=south west,inner sep=0] (image1) at (0,0) 
	     {\includegraphics[width=0.97\linewidth]
	        {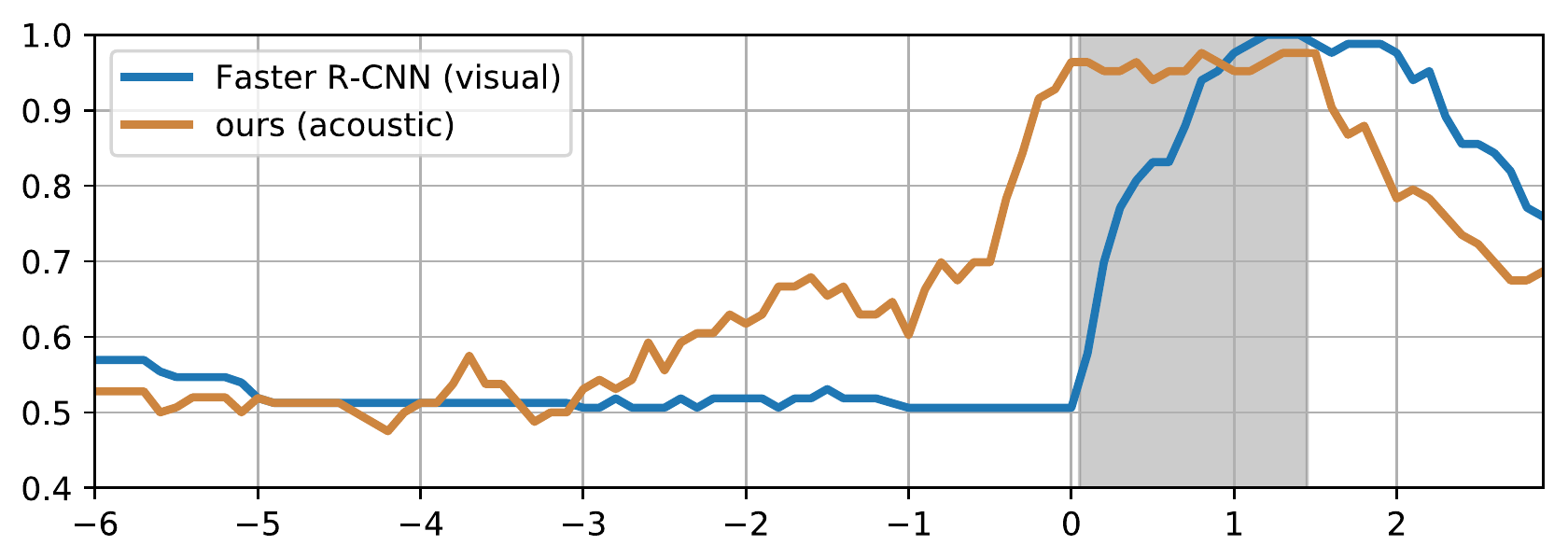}};
	     \node[anchor=south,below=of image1,inner sep=0,yshift=1.1cm] 
	        {$t_e-t_0$ [s]}; %
	     \node[inner sep=0,rotate=90,left=of image1,yshift=-0.9cm,xshift=1.05cm]
	        {Accuracy};
     \end{tikzpicture}
    \caption{Accuracy over test time $t_e$ of our acoustic and the visual baseline on 83 Static recordings.
        Gray region indicates the other vehicle is half-occluded and two labels, \classfront{} and
        either \classleft{} or \classright{}, are considered correct.}
    \label{fig:accuracy-vs-visual}
\end{figure}

\Updated{Figure \ref{fig:accuracy-vs-visual}} illustrates the accuracy on the test recordings for different evaluation times $t_e$.
The overlap region is indicated by the gray area after $t_e = t_0$ and its beginning thus marks when
a car enters the field of view.
At $t_e = t_0$, just before entering the view of the camera, the approaching car
can be detected with 0.94 accuracy by our method.
This accuracy is achieved more than one second ahead of the visual baseline,
showing that our acoustic detection gives the ego-vehicle additional reaction time.
\Updated{After $1.5$s a decreasing accuracy is reported,
since the leaving vehicle is not annotated
and only \classfront{} predictions are considered true positives.
The acoustic detector sometimes still predicts \classleft{}, or \classright{} once the car crossed over.
The Faster R-CNN accuracy also decreases: after $2$s the car is often completely occluded again.}

\begin{figure*}[t]
    \centering
    \subfloat[\classleft{} recordings]
          {\label{fig:predictivehorizonleft}
	       \singleplotxy{t}
	        {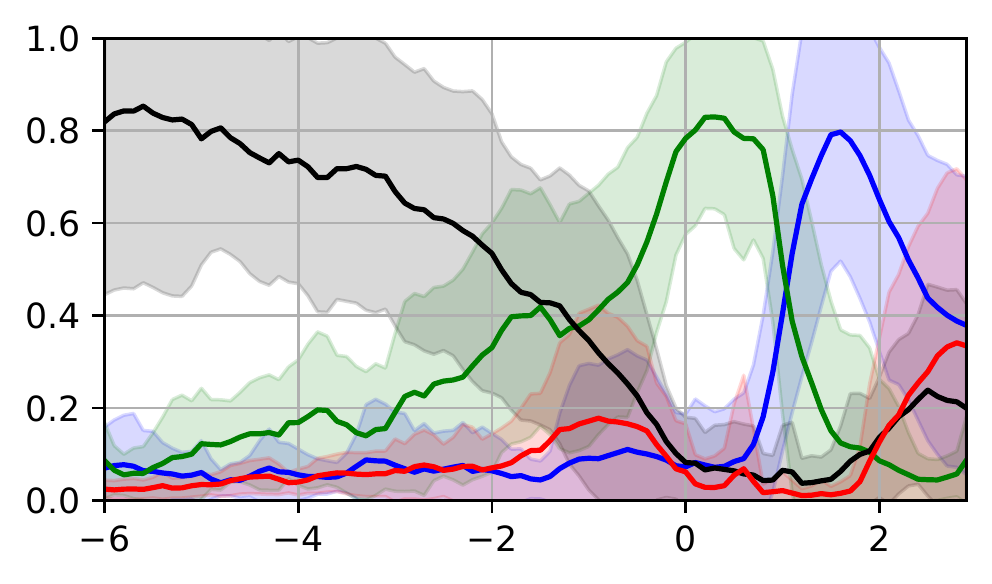}
          }
    \subfloat[\classright{} recordings]
          {\label{fig:predictivehorizonright}
	       \singleplotx{t}
	        {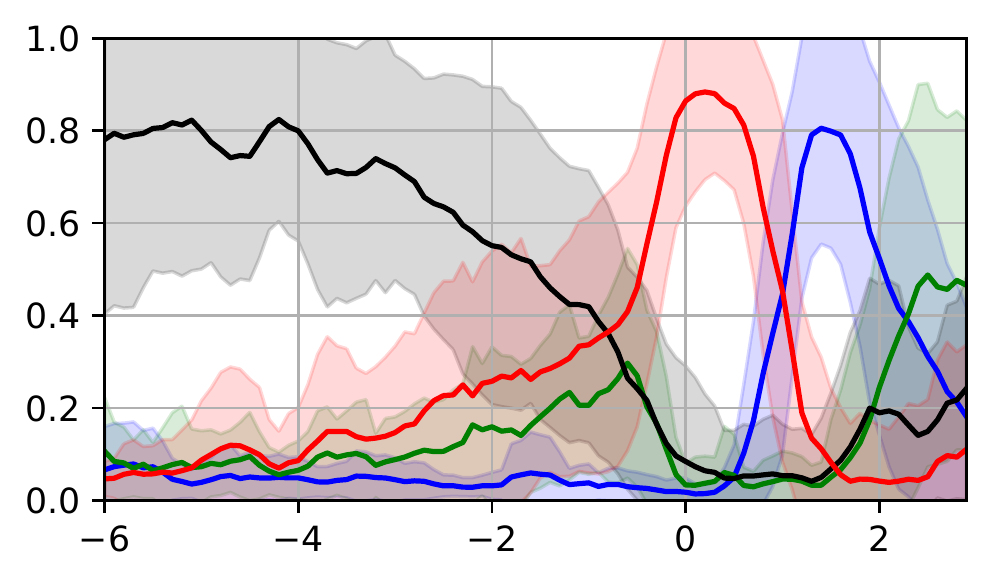}
          }
    \subfloat[\classneg{} recordings]
          {\label{fig:predictivehorizonnone}
	       \singleplotx{t}
	        {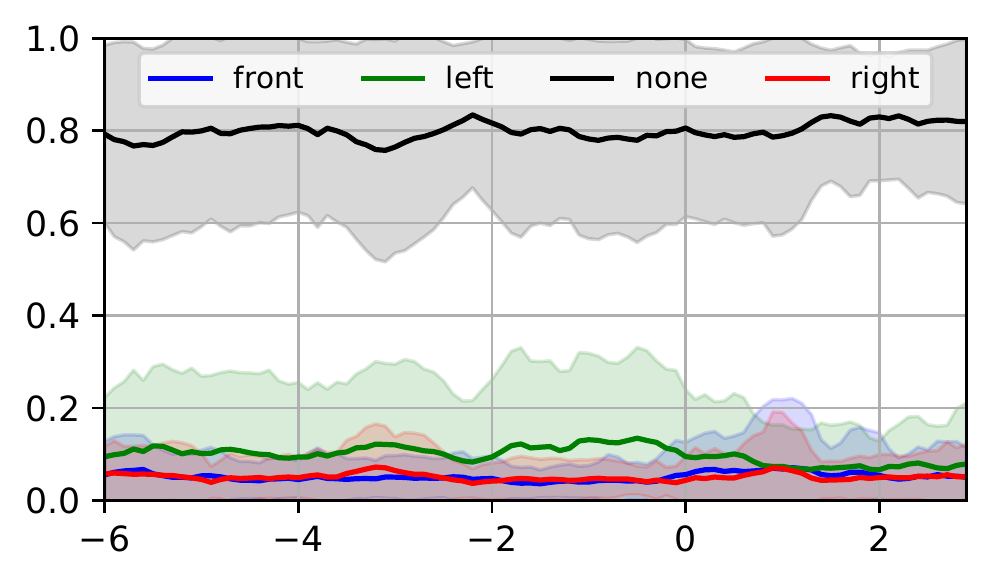}
          }
    \caption{
        Mean and std. dev. of predicted class probabilities at different times $t_e$ on 
        test set recordings of the Static data (blue is \classfront{}, green is \classleft{}, red is \classright{}, and black is \classneg{}).
        Each figure shows recordings of a different situation.
        The approaching vehicle appears in view just after
        $t_e - t_0 = 0$.
    }
    \label{fig:predictivehorizon}
\end{figure*}

Figure~\ref{fig:predictivehorizon} shows the per-class probabilities as a function of
extraction time $t_e$ on the test set, separated by recording situations.
The SVM class probabilities are obtained with the method in \cite{wu2004probability}.
The probabilities for \classleft{}
show that on average the model initially predicts that no car is approaching.
Towards $t_0$, the \classneg{} class becomes less likely and
the model increasingly favors the correct \classleft{} class.
A short time after $t_0$, the prediction flips to the \classfront{} class,
\Updated{and eventually switches to \classright{} as the car leaves line-of-sight}.
Similar (mirrored) behavior is observed for 
vehicles approaching from the right.
\Updated{The probabilities of \classleft{}/\classright{} rise until
the approaching vehicle is almost in line-of-sight,
which corresponds to the extraction time of the training samples.}
The \classneg{} class is constantly predicted as likeliest when no vehicle is approaching.
Overall, the prediction matches the events of the recorded situations remarkably well.


\begin{table}[h]
    \caption{Cross-validation results per environment on Dynamic data.
             }
    \label{tab:eval-within-loc-driving}
    \centering
        \begin{tabular}{c|c|cccc}
        Subset & Accuracy & $J_{\classleft{}}$ & $J_{\classfront{}}$ & $J_{\classright{}}$ & $J_{\classneg{}}$ \\
        \hline
        DAB & 0.76 & 0.41 & 0.80 & 0.44  & 0.65 \\
        DA  & 0.84 & 0.66 & 0.85 & 0.64 & 0.72 \\
        DB  & 0.75 & 0.33 & 0.81 & 0.42 & 0.64
        \end{tabular}
\end{table}

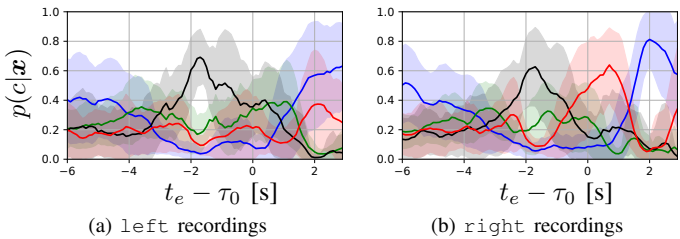
\begin{figure}[h]
    \hspace{-0.4cm}
    \subfloat[\classleft{} recordings]{
    \label{fig:predictivehorizonleftD}
         \begin{tikzpicture}
	     \node[anchor=south west,inner sep=0] (image2) at (0,0) 
	     {\includegraphics[width=0.232\textwidth]
	        {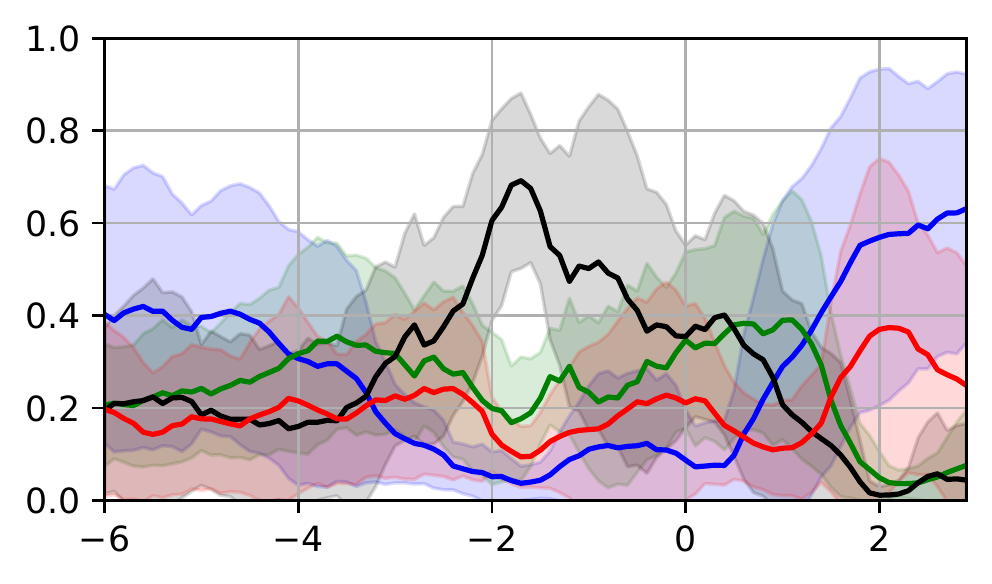}};
	     \node[inner sep=0] at (2.5,-0.1) {$t_e-\tau_0$ [s]};
	     \node[inner sep=0,rotate=90,left=of image2,yshift=-0.8cm,xshift=0.6cm]
	     {$p(c|\boldsymbol{x})$};
	     \end{tikzpicture}
     }
    \subfloat[\classright{} recordings]{
    \label{fig:predictivehorizonrightD}
         \begin{tikzpicture}
	     \node[anchor=south west,inner sep=0] (image2) at (0,0) 
	     {\includegraphics[width=0.232\textwidth]
            {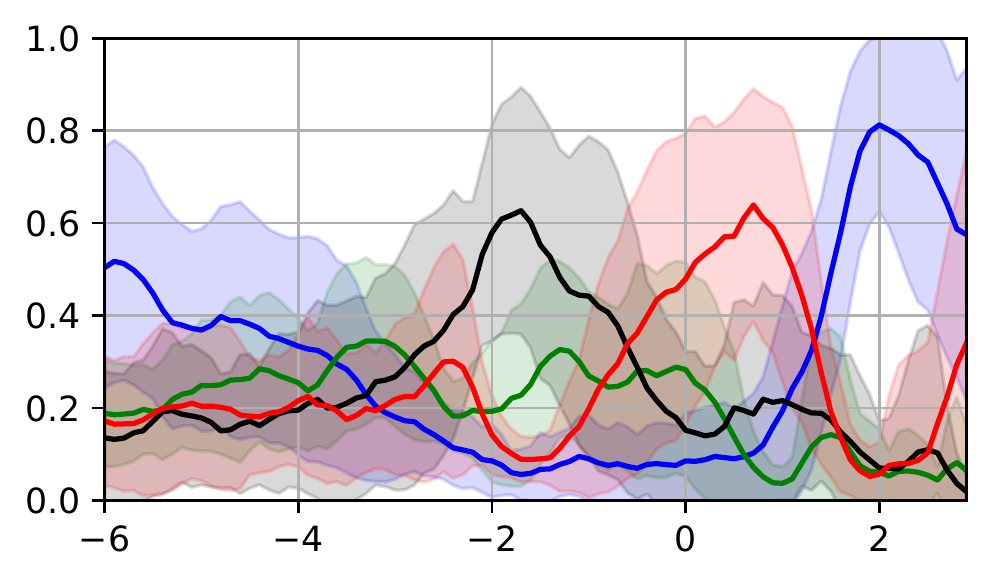}};
	     \node[inner sep=0] at (2.5,-0.1) {$t_e-\tau_0$ [s]};
	     \end{tikzpicture}
    }
    \caption{Mean and std. dev. of predicted class probabilities at different times $t_e$ on 
    \classleft{} and \classright{} test set recordings of the Dynamic data.
    The  ego-vehicle  reached the location  of  training  data  when
    $t_e - \tau_0 = 0.5 \mathrm{s}$.
             }
    \label{fig:predictivehorizonD}
\end{figure}

\subsection{Impact of the moving ego-vehicle}
\label{sec:dynamic-crossval}

\Updated{Next, our classifier is evaluated by cross-validation
per environment subset, as well as on the full Dynamic data.}
As for the Static data, 5-fold cross-validation is applied to each subset,
keeping the class distribution balanced across folds.

Table \ref{tab:eval-within-loc-driving} lists the corresponding metrics for each subset.
On the full Dynamic data (DAB), the accuracy indicates decent performance, but the metrics for
\classleft{} and \classright{} classes are much worse compared to 
the Static results in Table \ref{tab:eval-all-ablation}.
Separating subsets DA and DB reveals that the performance is highly dependent on the environment type.
In fact,
even with limited training data and large data variance from a driving ego-vehicle,
we obtain decent classification performance on type A environments,
and we notice that low \classleft{} and \classright{} performance mainly results from type B environments.
We hypothesize that the more confined type A environments reflect more target sounds
and are better shielded from potential noise sources.

We also analyze the temporal behavior of our method on Dynamic data.
Unfortunately, a fair comparison with a visual baseline
is not possible: the ego-vehicle often reaches the intersection early,
and the approaching vehicle is within line-of-sight but still outside the front-facing camera's field of view (cf. $\tau_0$ extraction in Section \ref{sec:dataset}).
Yet, the evolution of the predicted probabilities can be compared to those on the Static data in Section~\ref{sec:detection-time}.
Figure \ref{fig:predictivehorizonD} illustrates the average predicted probabilities
over 59 Dynamic test set recordings from all locations,
after training on samples from the remaining 233 recordings.
The classifier on average correctly predicts \classright{} samples
(Figure \ref{fig:predictivehorizonrightD}), 
between $t_e = \tau_0$ to $t_e = \tau_0 + 0.5 \mathrm{s}$.
Of the \classleft{} recordings at these times, many are falsely predicted
as \classneg{}, only few are confused with \classright{}.
Furthermore, the changing ego-perspective of the vehicle results in alternating
DoA-energy \Updated{directions} and thus class predictions, compared to the Static results in Figure \ref{fig:predictivehorizon}.
This indicates that it might help to include the ego-vehicle's relative position as
an additional feature, and obtain more varied training data to cover the positional variations.

\subsection{Generalization across acoustic environments}

We here study how the performance is affected when the classifier is trained on all samples from one environment type and
evaluated on all samples of the other type.
In Table~\ref{tab:eval-across-env}, combinations of training and test
sets are listed.
Compared to the results for Static and Dynamic data
(see Tables \ref{tab:eval-all-ablation} and \ref{tab:eval-within-loc-driving}),
the reported results in the table show a general trend:
If the classifier is trained on one environment and tested \Updated{on the other},
it performs worse than when samples of the same location are \Updated{used}.
In particular, the classifier trained on SB and tested on SA is not
\Updated{able} to correctly classify samples of \classleft{} and \classright{}
while inverse training and testing performs much better.
On the Dynamic data, such pronounced effects are not visible, but overall the accuracy decreases compared to the Static data.
In summary, the \Updated{reflection patterns vary from} one environment to another,
\Updated{yet at some locations the patterns appear more distinct and robust than those at others}.

\begin{table}[h]
    \caption{Generalization across locations and environments.
             }
    \label{tab:eval-across-loc}
    \label{tab:eval-across-env}
    \centering
        \begin{tabular}{cc|c|cccc}
        Training & Test & Accuracy & $J_{\classleft{}}$ & $J_{\classfront{}}$ & $J_{\classright{}}$ & $J_{\classneg{}}$ \\
        \hline
         SB  & SA  & 0.66 & 0.03 & 0.66 & 0.03 & 0.62 \\
         SA  & SB  & 0.79 & 0.42 & 0.82 & 0.61 & 0.67 \\
        \hline
        DB   & DA  & 0.53 & 0.16 & 0.70 & 0.25 & 0.16 \\
        DA   & DB  & 0.56 & 0.21 & 0.50 & 0.29 & 0.46
        \end{tabular}
\end{table}

\subsection{Microphone array configuration}
\label{sec:number-of-mics}
\Updated{%
Our array with 56 microphones enables evaluation of different spatial configurations with $\nummics{} < 56$.
For various subsets of $\nummics{}$ microphones, we randomly sample 100 out of $\binom{56}{\nummics{}}$ possible microphone configurations, and cross-validate on the Static data.
Interestingly, the best configuration with $\nummics{} = 7$ already achieves similar accuracy as with $\nummics{} = 56$.
With $\nummics{} = 2/3$ the accuracy is already $0.82/0.89$,
but with worse performance on \classleft{} and \classright{}.
Large variance between samples highlights the importance of a thorough search of spatial configurations.
Reducing $\nummics{}$ also leads to faster inference time, specifically 0.24/0.14/0.04s for $\nummics{}=56/28/14$ using our unoptimized implementation}.

%% file: tex/conclusions.tex
We \Updated{showed} that a vehicle mounted microphone array can be used to acoustically detect 
approaching vehicles behind blind corners
\Updated{from their wall reflections.
In our experimental setup, our method achieved an accuracy of 0.92
on the 4-class hidden car classification task for a static ego-vehicle,
and up to 0.84 in some environments while driving.
An approaching vehicle was detected with the same accuracy as our visual
baseline already more than one second ahead,
a crucial advantage in such critical situations.} %

\Updated{While these initial findings are encouraging, our results have several limitations.
The experiments included only few locations and few different oncoming vehicles,
and while our method performed well on one environment, it had difficulties on the other,
and did not perform reliably in unseen test environments.
To expand the applicability, we expect that more representative data is needed
to capture a broad variety of environments, vehicle positions and velocities,
and the presence of multiple sound sources.
Rather than generalizing across environments, additional input from map data or other sensor measurements
could help to discriminate acoustic environments and to classify the reflection patterns accordingly.
More data also enables end-to-end learning of low-level features,
potentially capturing cues our DoA-based approach currently ignores (e.g. Doppler, sound volume),
and perform multi-source detection and classification in one pass~\cite{he2018deep}.
Ideally a suitable self-supervised learning scheme is developed~\cite{gan19},
though a key challenge is that actual occluded sources cannot immediately be visually detected.
}

%% file: root.bbl
\begin{thebibliography}{10}
\providecommand{\url}[1]{#1}
\csname url@rmstyle\endcsname
\providecommand{\newblock}{\relax}
\providecommand{\bibinfo}[2]{#2}
\providecommand\BIBentrySTDinterwordspacing{\spaceskip=0pt\relax}
\providecommand\BIBentryALTinterwordstretchfactor{4}
\providecommand\BIBentryALTinterwordspacing{\spaceskip=\fontdimen2\font plus
\BIBentryALTinterwordstretchfactor\fontdimen3\font minus
  \fontdimen4\font\relax}
\providecommand\BIBforeignlanguage[2]{{%
\expandafter\ifx\csname l@#1\endcsname\relax
\typeout{** WARNING: IEEEtran.bst: No hyphenation pattern has been}%
\typeout{** loaded for the language `#1'. Using the pattern for}%
\typeout{** the default language instead.}%
\else
\language=\csname l@#1\endcsname
\fi
#2}}

\bibitem{keller2011active}
C.~G. Keller, T.~Dang, H.~Fritz, A.~Joos, C.~Rabe, and D.~M. Gavrila, ``Active
  pedestrian safety by automatic braking and evasive steering,'' \emph{IEEE
  T-ITS}, vol.~12, no.~4, pp. 1292--1304, 2011.

\bibitem{machardy2018v2x}
Z.~MacHardy, A.~Khan, K.~Obana, and S.~Iwashina, ``{V2X} access technologies:
  Regulation, research, and remaining challenges,'' \emph{IEEE Comm. Surveys \&
  Tutorials}, vol.~20, no.~3, pp. 1858--1877, 2018.

\bibitem{argentieri2015survey}
S.~Argentieri, P.~Danes, and P.~Sou{\`e}res, ``A survey on sound source
  localization in robotics: From binaural to array processing methods,''
  \emph{Computer Speech \& Language}, vol.~34, no.~1, pp. 87--112, 2015.

\bibitem{rascon2017localization}
C.~Rascon and I.~Meza, ``Localization of sound sources in robotics: A review,''
  \emph{Robotics \& Autonomous Systems}, vol.~96, pp. 184--210, 2017.

\bibitem{wang2018acoustic}
L.~Wang and A.~Cavallaro, ``Acoustic sensing from a multi-rotor drone,''
  \emph{IEEE Sensors Journal}, vol.~18, no.~11, pp. 4570--4582, 2018.

\bibitem{lindell2019acoustic}
D.~B. Lindell, G.~Wetzstein, and V.~Koltun, ``Acoustic non-line-of-sight
  imaging,'' in \emph{Proc. of IEEE CVPR}, 2019, pp. 6780--6789.

\bibitem{okutani2012outdoor}
K.~Okutani, T.~Yoshida, K.~Nakamura, and K.~Nakadai, ``Outdoor auditory scene
  analysis using a moving microphone array embedded in a quadrocopter,'' in
  \emph{IEEE/RSJ IROS}.\hskip 1em plus 0.5em minus 0.4em\relax IEEE, 2012, pp.
  3288--3293.

\bibitem{an2018reflection}
I.~An, M.~Son, D.~Manocha, and S.-e. Yoon, ``Reflection-aware sound source
  localization,'' in \emph{ICRA}.\hskip 1em plus 0.5em minus 0.4em\relax IEEE,
  2018, pp. 66--73.

\bibitem{jang2015development}
Y.~Jang, J.~Kim, and J.~Kim, ``The development of the vehicle sound source
  localization system,'' in \emph{APSIPA}.\hskip 1em plus 0.5em minus
  0.4em\relax IEEE, 2015, pp. 1241--1244.

\bibitem{stelling2015traffic}
A.~Stelling-Ko{\'n}czak, M.~Hagenzieker, and B.~V. Wee, ``Traffic sounds and
  cycling safety: The use of electronic devices by cyclists and the quietness
  of hybrid and electric cars,'' \emph{Transport Reviews}, vol.~35, no.~4, pp.
  422--444, 2015.

\bibitem{mizumachi2014robust}
M.~Mizumachi, A.~Kaminuma, N.~Ono, and S.~Ando, ``Robust sensing of approaching
  vehicles relying on acoustic cues,'' \emph{Sensors}, vol.~14, no.~6, pp.
  9546--9561, 2014.

\bibitem{padmanabhan2014acoustics}
A.~V. Padmanabhan, H.~Ravichandran, \emph{et~al.}, ``Acoustics based vehicle
  environmental information,'' SAE, Tech. Rep., 2014.

\bibitem{asahi2011development}
K.~Asahi, H.~Banno, O.~Yamamoto, A.~Ogawa, and K.~Yamada, ``Development and
  evaluation of a scheme for detecting multiple approaching vehicles through
  acoustic sensing,'' in \emph{IV Symposium}.\hskip 1em plus 0.5em minus
  0.4em\relax IEEE, 2011, pp. 119--123.

\bibitem{singh2012non}
V.~Singh, K.~E. Knisely, \emph{et~al.}, ``Non-line-of-sight sound source
  localization using matched-field processing,'' \emph{J. of the Acoustical
  Society of America}, vol. 131, no.~1, pp. 292--302, 2012.

\bibitem{toyoda2014traffic}
T.~Toyoda, N.~Ono, S.~Miyabe, T.~Yamada, and S.~Makino, ``Traffic monitoring
  with ad-hoc microphone array,'' in \emph{Int. Workshop on Acoustic Signal
  Enhancement}.\hskip 1em plus 0.5em minus 0.4em\relax IEEE, 2014, pp.
  318--322.

\bibitem{ishida2018saved}
S.~Ishida, J.~Kajimura, M.~Uchino, S.~Tagashira, and A.~Fukuda, ``{SAVeD}:
  Acoustic vehicle detector with speed estimation capable of sequential vehicle
  detection,'' in \emph{ITSC}.\hskip 1em plus 0.5em minus 0.4em\relax IEEE,
  2018, pp. 906--912.

\bibitem{sandberg2010vehicles}
U.~Sandberg, L.~Goubert, and P.~Mioduszewski, ``Are vehicles driven in electric
  mode so quiet that they need acoustic warning signals,'' in \emph{Int.
  Congress on Acoustics}, 2010.

\bibitem{iversen2015measurement}
L.~M. Iversen and R.~S.~H. Skov, ``Measurement of noise from electrical
  vehicles and internal combustion engine vehicles under urban driving
  conditions,'' \emph{Euronoise}, 2015.

\bibitem{robart2013evader}
R.~Robart, E.~Parizet, J.-C. Chamard, \emph{et~al.}, ``{eVADER}: A perceptual
  approach to finding minimum warning sound requirements for quiet cars.'' in
  \emph{AIA-DAGA 2013 Conference on Acoustics}, 2013.

\bibitem{lee2017objective}
S.~K. Lee, S.~M. Lee, T.~Shin, and M.~Han, ``Objective evaluation of the sound
  quality of the warning sound of electric vehicles with a consideration of the
  masking effect: Annoyance and detectability,'' \emph{Int. Journal of
  Automotive Tech.}, vol.~18, no.~4, pp. 699--705, 2017.

\bibitem{scheibler2018pyroomacoustics}
R.~Scheibler, E.~Bezzam, and I.~Dokmani{\'c}, ``Pyroomacoustics: A python
  package for audio room simulation and array processing algorithms,'' in
  \emph{ICASSP}.\hskip 1em plus 0.5em minus 0.4em\relax IEEE, 2018, pp.
  351--355.

\bibitem{dibiase2000high}
J.~H. DiBiase, \emph{A high-accuracy, low-latency technique for talker
  localization in reverberant environments using microphone arrays}.\hskip 1em
  plus 0.5em minus 0.4em\relax Brown University Providence, RI, 2000.

\bibitem{hornikx2011modelling}
M.~Hornikx and J.~Forss{\'e}n, ``Modelling of sound propagation to
  three-dimensional urban courtyards using the extended {F}ourier pstd
  method,'' \emph{Applied Acoustics}, vol.~72, no.~9, pp. 665--676, 2011.

\bibitem{zhang2017surround}
W.~Zhang, P.~N. Samarasinghe, H.~Chen, and T.~D. Abhayapala, ``Surround by
  sound: A review of spatial audio recording and reproduction,'' \emph{Applied
  Sciences}, vol.~7, no.~5, p. 532, 2017.

\bibitem{osako2017supervised}
K.~Osako, Y.~Mitsufuji, \emph{et~al.}, ``Supervised monaural source separation
  based on autoencoders,'' in \emph{ICASSP}.\hskip 1em plus 0.5em minus
  0.4em\relax IEEE, 2017, pp. 11--15.

\bibitem{saxena2009learning}
A.~Saxena and A.~Y. Ng, ``Learning sound location from a single microphone,''
  in \emph{ICRA}.\hskip 1em plus 0.5em minus 0.4em\relax IEEE, 2009, pp.
  1737--1742.

\bibitem{salamon2017deep}
J.~Salamon and J.~P. Bello, ``Deep convolutional neural networks and data
  augmentation for environmental sound classification,'' \emph{IEEE Signal
  Processing Letters}, vol.~24, no.~3, pp. 279--283, 2017.

\bibitem{valada2018deep}
A.~Valada, L.~Spinello, and W.~Burgard, ``Deep feature learning for
  acoustics-based terrain classification,'' in \emph{Robotics Research}.\hskip
  1em plus 0.5em minus 0.4em\relax Springer, 2018, pp. 21--37.

\bibitem{yalta2017}
N.~Yalta, K.~Nakadai, and T.~Ogata, ``Sound source localization using deep
  learning models,'' \emph{J. of Robotics and Mechatronics}, vol.~29, no.~1,
  pp. 37--48, 2017.

\bibitem{he2018deep}
W.~He, P.~Motlicek, and J.-M. Odobez, ``Deep neural networks for multiple
  speaker detection and localization,'' in \emph{ICRA}.\hskip 1em plus 0.5em
  minus 0.4em\relax IEEE, 2018, pp. 74--79.

\bibitem{gan19}
C.~Gan, H.~Zhao, P.~Chen, D.~Cox, and A.~Torralba, ``Self-supervised moving
  vehicle tracking with stereo sound,'' in \emph{Proc. of ICCV}, 2019.

\bibitem{scheiner2020seeing}
N.~Scheiner, F.~Kraus, F.~Wei, \emph{et~al.}, ``Seeing around street corners:
  Non-line-of-sight detection and tracking in-the-wild using doppler radar,''
  in \emph{Proc. of IEEE CVPR}, 2020, pp. 2068--2077.

\bibitem{ren2015faster}
S.~Ren, K.~He, R.~Girshick, and J.~Sun, ``Faster r-cnn: Towards real-time
  object detection with region proposal networks,'' in \emph{Advances in neural
  information processing systems}, 2015, pp. 91--99.

\bibitem{ferranti2019safevru}
L.~Ferranti, B.~Brito, E.~Pool, Y.~Zheng, \emph{et~al.}, ``{SafeVRU}: A
  research platform for the interaction of self-driving vehicles with
  vulnerable road users,'' in \emph{IV Symposium}.\hskip 1em plus 0.5em minus
  0.4em\relax IEEE, 2019, pp. 1660--1666.

\bibitem{pedregosa2011scikit}
F.~Pedregosa, G.~Varoquaux, \emph{et~al.}, ``Scikit-learn: Machine learning in
  python,'' \emph{JMLR}, vol.~12, no. Oct, pp. 2825--2830, 2011.

\bibitem{sarradj2017acoular}
E.~Sarradj and G.~Herold, ``A python framework for microphone array data
  processing,'' \emph{Applied Acoustics}, vol. 116, pp. 50--58, 2017.

\bibitem{detectron2}
Y.~Wu, A.~Kirillov, F.~Massa, W.-Y. Lo, and R.~Girshick, ``Detectron2,''
  \url{https://github.com/facebookresearch/detectron2}, 2019.

\bibitem{wu2004probability}
T.-F. Wu, C.-J. Lin, and R.~C. Weng, ``Probability estimates for multi-class
  classification by pairwise coupling,'' \emph{JMLR}, vol.~5, no. Aug, pp.
  975--1005, 2004.

\end{thebibliography}
